\documentclass[journal]{IEEEtran}
\ifCLASSINFOpdf
\else
\fi
\usepackage{url}
\usepackage{cite}
\usepackage{amsmath,amssymb,amsfonts}
\usepackage{mathtools}
\usepackage{bbm}

\usepackage{changes}

\usepackage{textcomp}
\usepackage{multicol}
\usepackage{multirow}
\usepackage{algpseudocode}
\usepackage[ruled,vlined]{algorithm2e}
\usepackage{tabularx}
\usepackage{adjustbox}  
\usepackage{dblfloatfix}
\usepackage[para]{threeparttable}
\usepackage{caption}
\usepackage{stackengine}
\usepackage{comment}
\usepackage{xcolor}
\usepackage{subfig}
\usepackage{graphicx}
\usepackage{fancyhdr}

\SetKwInput{KwInput}{Input}                
\SetKwInput{KwOutput}{Output}              

\def\x{{\mathbf x}}

\def\sgn{{\rm sgn}}

\def\x{{\bf x}}

\def\K{{\bf K}}

\def\x{{\mathbf x}}

\def\w{{\bf w}}

\begin{document}
%
\title{Multiplierless MP-Kernel Machine For Energy-efficient Edge Devices}
%
%
%

\author{Abhishek~Ramdas~Nair$^*$,
        Pallab~Kumar~Nath$^*$,
        Shantanu~Chakrabartty,
        and~Chetan~Singh~Thakur,
\thanks{Abhishek Ramdas Nair, Pallab Kumar Nath and Chetan Singh Thakur are with the Department
of Electronic Systems Engineering, Indian Institute of Science, Bangalore,
KA, 560012 INDIA e-mail: (abhisheknair@iisc.ac.in, pkniitkgp@gmail.com, csthakur@iisc.ac.in).}
\thanks{Shantanu Chakrabartty is with Department of Electrical and Systems Engineering, Washington University in St. Louis,USA, 63130. e-mail:shantanu@wustl.edu}
\thanks{$^*$ Both Abhishek Ramdas Nair and Pallab Kumar Nath have contributed equally to this paper.}%
\thanks}

%
%

\markboth{}%
{Nair \MakeLowercase{\textit{et al.}}: Multiplierless MP-Kernel Machine For Energy-efficient Edge Devices}
%



\maketitle

\begin{abstract}
We present a novel framework for designing multiplierless kernel machines that can be used on resource-constrained platforms like intelligent edge devices. The framework uses a piecewise linear (PWL) approximation based on a margin propagation (MP) technique and uses only addition/subtraction, shift, comparison, and register underflow/overflow operations. We propose a hardware-friendly MP-based inference and online training algorithm that has been optimized for a Field Programmable Gate Array (FPGA) platform. Our FPGA implementation eliminates the need for DSP units and reduces the number of LUTs. By reusing the same hardware for inference and training, we show that the platform can overcome classification errors and local minima artifacts that result from the MP approximation. The implementation of this proposed multiplierless MP-kernel machine on FPGA results in an estimated energy consumption of 13.4 pJ and power consumption of 107 mW with \texttildelow 9k LUTs and FFs each for a $256 \times 32$ sized kernel making it superior in terms of power, performance, and area compared to other comparable implementations.
\end{abstract}


%
\IEEEpeerreviewmaketitle

\section{Introduction}
%
%
%
%
\IEEEPARstart{E}{dge} computing is transforming the way data is being handled, processed, and delivered in various applications\cite{satyanarayanan2017emergence}\cite{shi2016edge}. At the core of edge computing platforms are embedded machine learning (ML) 
algorithms that make decisions in real-time, which endows these platforms with greater autonomy~\cite{wang2018edge}\cite{zhu2020toward}. 
Common edge-ML architectures reported in the literature are based on a deep neural network~\cite{li2018learning} or support vector machines~\cite{flouri2006training}, and one of the active areas of research is to be able to improve the energy efficiency of these ML architectures, both for inference and learning~\cite{cui2019stochastic}. To achieve this, the hardware models are first trained offline, and the trained models are then optimized for 
energy efficiency using pruning or sparsification techniques~\cite{han2015deep} before being deployed on the edge platform.

\begin{figure}[ht]
    \centering{\includegraphics[page=1,scale=0.32,trim=4 4 4 4,clip]{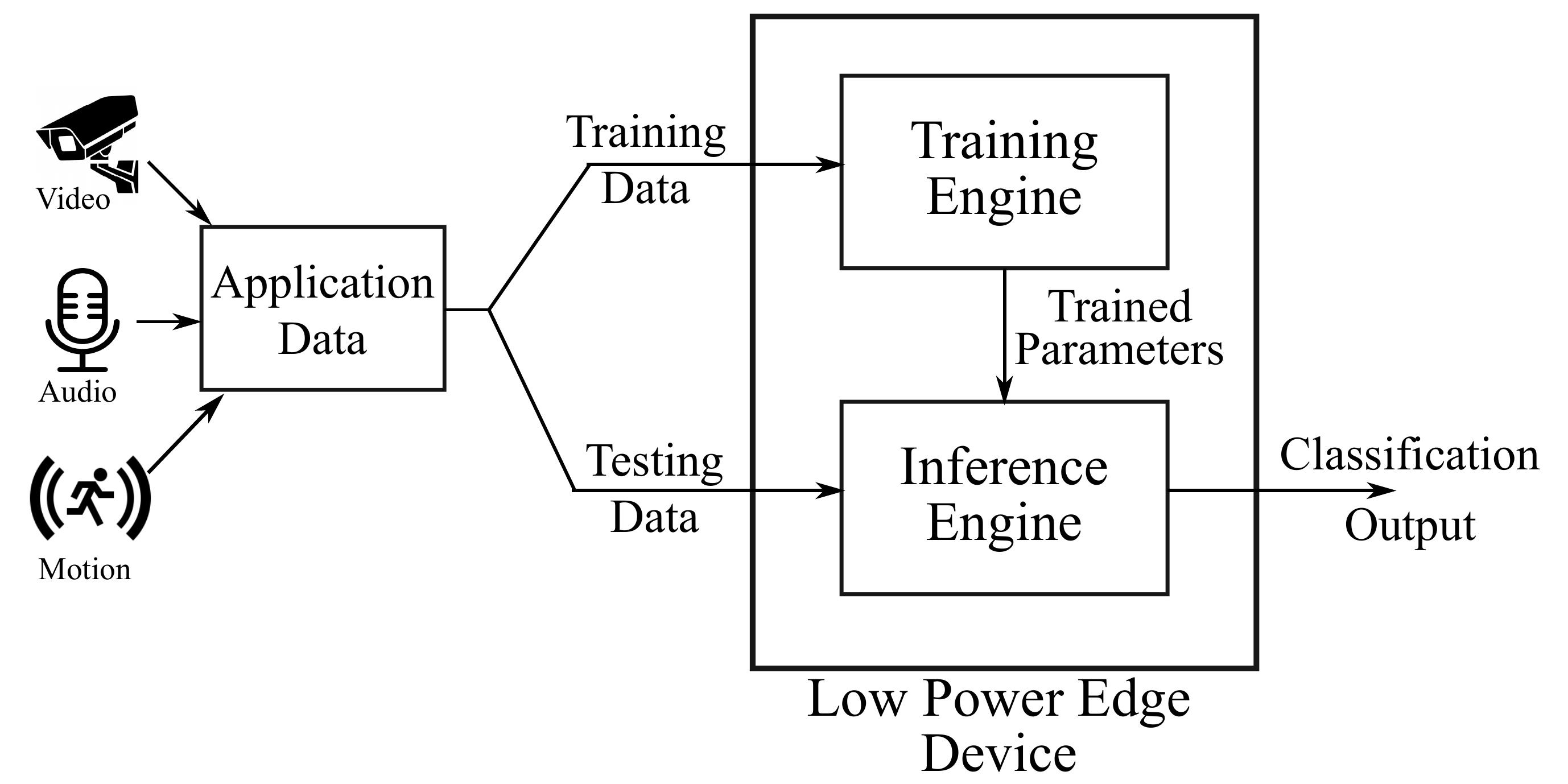}}
    \caption{Edge Device with Online Learning Capability}
    \label{Edge_Device}
\end{figure}

An example of such design-flows is the binary or the quaternary neural networks, which are compressed and energy-efficient variants of deep-neural network inference engines \cite{courbariaux2015binaryconnect} \cite{rastegari2016xnor}. However, robust training of quantized deep learning architectures still requires full precision training \cite{hubara2017quantized}. Also, deep neural networks require a large amount of training data ~\cite{heydari2018effect} which is generally unavailable for many applications. On the other hand, Support Vector Machines (SVMs) can provide good classification results with significantly less training data. The convex nature of SVM optimization makes its training more robust and more interpretable~\cite{cortes1995support}. 
SVMs have also been shown to outperform deep learning systems for detecting rare  events\cite{tang2008svms}\cite{palaniappan2012abnormal}, which is generally the case for many IoT applications. One such IoT-based edge device architecture is depicted in Fig.~\ref{Edge_Device}. Data from video surveillance, auditory event, or motion sensor can be analyzed, and the system can be trained on the device to produce robust classification models.

At a fundamental level, SVMs and other ML architectures extensively use Matrix-Vector-Multiplication (MVM) operations. One way to improve the overall system energy efficiency is to reduce the complexity or minimize MVM operations. In literature, many approximations and
reduced precision MVM techniques have been proposed~\cite{choi2019approximate} and have produced improvements in energy efficiency without significantly
sacrificing classification accuracy. Kernel machines have similar inference engine as SVMs, sharing the execution characteristics with SVM~\cite{doi:10.1198/jasa.2003.s270}. This paper proposes a kernel machine architecture that eliminates the need for multipliers. Instead, it uses more fundamental but energy efficient and optimal computational primitives like addition/subtraction, shift, and overflow/underflow operations.
For example, in a 45 nm CMOS technology, it has been shown that an 8-bit multiplication operation consumes 0.2 pJ of energy, whereas an 8-bit addition/subtraction operation consumes only 0.03 pJ of energy\cite{horowitz20141}. Shift and comparison operations consume even less energy than additions and subtractions, and
underflow/overflow operations do not consume any additional energy at all. 
To achieve this multiplierless mapping, the proposed architecture uses a margin propagation (MP) approximation technique originally proposed for analog computing~\cite{chakrabartty2004margin}. MP approximation has been optimized for a digital edge computing platform like field-programmable gate arrays (FPGA) in this work. We show that for inference, the MP approximation is computed as a part of learning, and all
the computational steps can be pipelined and parallelized for other MP approximation operations. In addition to reporting an MP approximation-based inference engine, we also report an online training algorithm that uses a gradient-descent technique in conjunction with hyper-parameter annealing. We show that the same hardware can be reused for both training and inference for the proposed MP kernel machine. As a result, the MP approximation errors can be effectively reduced. Since kernel machine inference and SVM inference have similar equations, we compare our system with traditional SVMs. We show that MP-based kernel machines can achieve similar classification accuracy as floating-point SVMs without using multipliers or equivalently any MVMs.

The main contributions of this paper are as follows:
\begin{itemize}
  \item Design and optimization of MP-approximation using iterative binary addition, shift, comparison, and underflow/overflow operations.
  \item Implementation of energy-efficient MP-based inference on an FPGA-based edge platform with multiplierless architecture that eliminates the need for DSP units.
  \item Online training of MP-based kernel machine that reuses the inference hardware. 
\end{itemize}

 The rest of this paper is organized as follows. Section II briefly discusses related work, followed by section III, where we explain the concept of multiplierless inner product computation and the related MP-based approximation. Section \ref{MP_Math} presents the kernel machine formulation based on MP theory. Section \ref{On-chip} details the online training of the system. Section \ref{FPGA} provides the FPGA implementation details and contrasts with other hardware implementations of MP-based kernel machine algorithm.
Section \ref{Results} discusses results obtained with few classification datasets and compares our multiplierless system with other SVM implementations, and Section \ref{Conclusion} concludes this paper and discusses potential use cases.

\section{Related Work} \label{related_work}

Energy-efficient SVM implementations have been reported for both digital~\cite{genov2003kerneltron} and analog hardware~\cite{chakrabartty2003forward}, which also exploit the inherent parallelism and regularity in MVM computation. In \cite{kang2009chip}, an analog circuit architecture of Gaussian-kernel SVM having on-chip training capability has been developed. Even though it has a scalable processor configuration, the circuit size increases in proportion to the number of learning samples. Such designs are not scalable as we increase the dataset size for edge devices due to hardware constraints. Analog domain architectures tend to have lower classification latency \cite{chakrabartty2007sub}, but they support simple classification models with small feature dimensions. In \cite{kyrkou2011parallel}, an SVM architecture has been reported using an array of processing elements in a systolic chain configuration implemented on an FPGA. By exploiting resource sharing capability among the processing units and efficient memory management, such an architecture permitted a higher throughput and a more scalable design.  

Digital and optimized FPGA implementation of SVM has been reported in~\cite{kyrkou2013embedded} using a cascaded architecture. A hardware reduction method is used to reduce the overheads due to the implementation of additional stages in the cascade, leading to significant resource and power savings for embedded applications. The use of cascaded SVMs increase the classification speeds, but the improved performance comes at the cost of additional hardware resources.

Ideally, a single-layered SVM should be enough for classification at the edge. In \cite{jiang2017fpga}, SVM prediction on an FPGA platform has been explored for ultrasonic flaw detection. A similar system has been implemented in \cite{martins2019svm} using Hamming distance for onboard classification of hyperspectral images. Since the SVM training phase needs a large amount of computation power and memory space, and often retraining is not required, the SVM training was realized offline. Therefore, the authors chose to accelerate only the classifier's decision function.
 In \cite{han2020low}, the authors use stochastic computing as a low hardware cost and low power consumption SVM classifier implementation for real-time EEG based gesture recognition. Even though, this novel design has merits of energy efficient inference, the training framework had to be handled offline on a power hungry system.

Often, training offline is not ideal for edge devices, primarily when the device is operating in dynamic environments, i.e., ever-changing parameters. In such cases, retraining of the ML architecture becomes important. One such system is discussed in \cite{kuan2011vlsi}, where sequential minimal optimization learning algorithm is implemented as an IP on FPGA. This can be leveraged for online learning systems. However, this system standalone cannot provide an end-to-end training and inference system. Another online training capable system was reported in \cite{dey2018highly} and used an FPGA implementation of a sparse neural network capable of online training and inference. The platform could be reconfigured to trade-off resource utilization with training time while keeping the network architecture the same. Even though the same platform can be used for training and inference, memory management and varying resource utilization based on training time make it less conducive to deploy on the edge device. Reconfiguration of the device would require additional usage of a microcontroller, increasing the system's overall power consumption.

As hardware complexity and power reduction are a major concern in these designs, the authors in \cite{mandal2014implementation} implement a multiplierless SVM kernel for classification in hardware instead of using a conventional vector product kernel. The data-flow amongst processing elements is handled using a parallel pipelined systolic array system. In the multiplierless block the authors use Canonic Signed Digit (CSD) to reduce the maximum number of adders. CSD is a number system by which a floating-point number can be represented in two’s complement form. The representation uses only -1, 0, +1 (or -, 0, +) symbols with each position denoting the addition and subtraction of power of 2. Despite being multiplierless, the system consumes many Digital Signal Processors (DSPs) due to the usage of  polynomial kernel having exponent operation. The usage of DSPs increases the power consumption of the design.

Another system that uses multiplierless online training was reported in \cite{xue2020real}. The authors use a logarithm function based on a look-up table, and a float-to-fixed point transform to simplify the calculations in a Naive Bayes classification algorithm. A novel format of a logarithm look-up table and shift operations are used to replace multiplication and division operations. Such systems incur an overhead of generating the logarithmic look-up table for most operations. The authors chose to calculate logarithmic values offline and store them in memory, contributing to increased memory access for every operation. 

\captionsetup[subfigure]{labelformat=empty}
\begin{figure*}[ht]
    \centering
    \subfloat[]{\label{fig:1Dwx}\stackinset{l}{.1in}{t}{.1in}{(a)}{\includegraphics[page=1,scale=0.2,trim=330 0 300 0,clip]{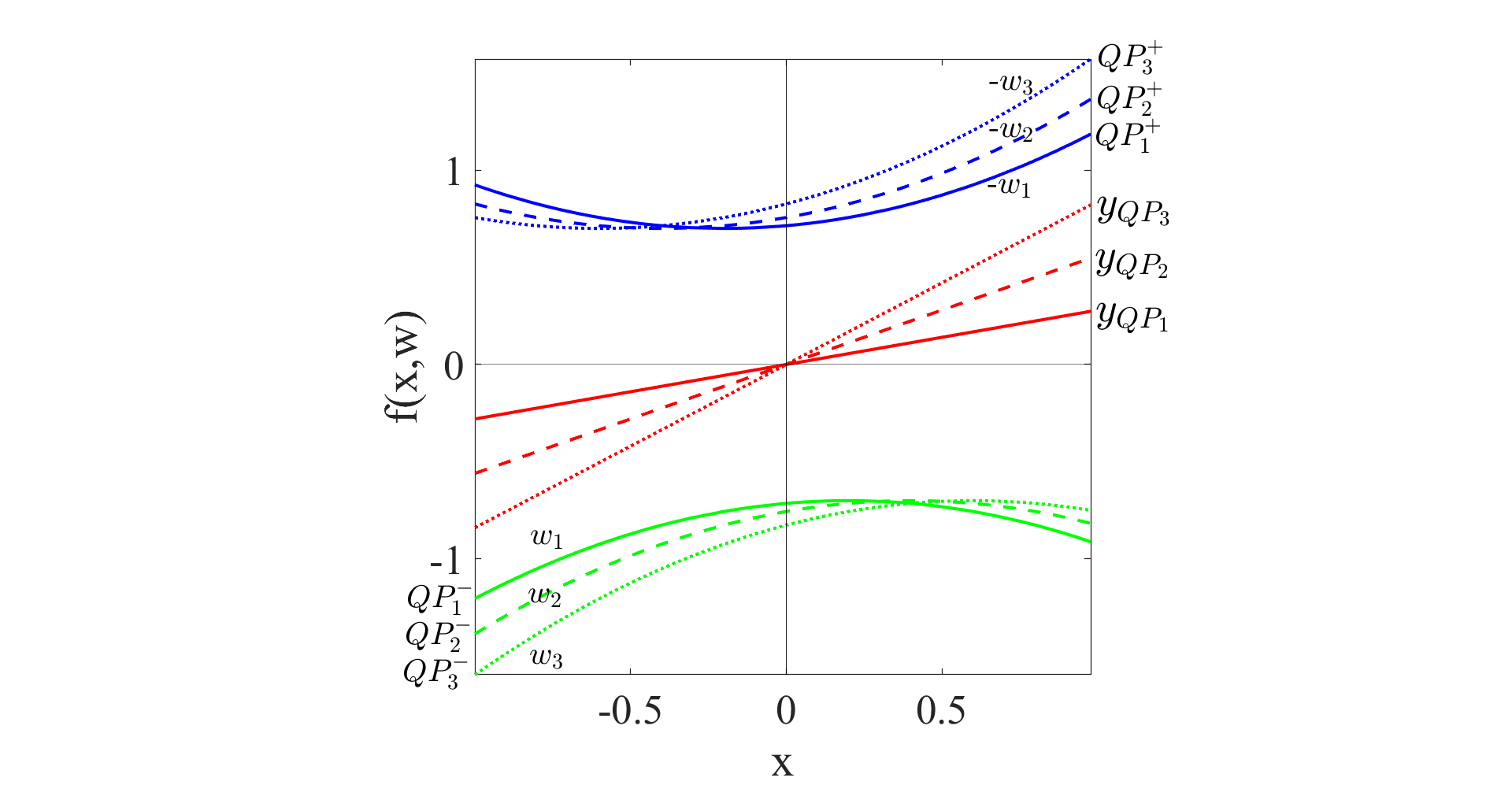}}}
    \subfloat[]{\label{fig:1Dlogewx}\stackinset{l}{.1in}{t}{.1in}{(b)}{\includegraphics[page=1,scale=0.2,trim=330 0 300 0,clip]{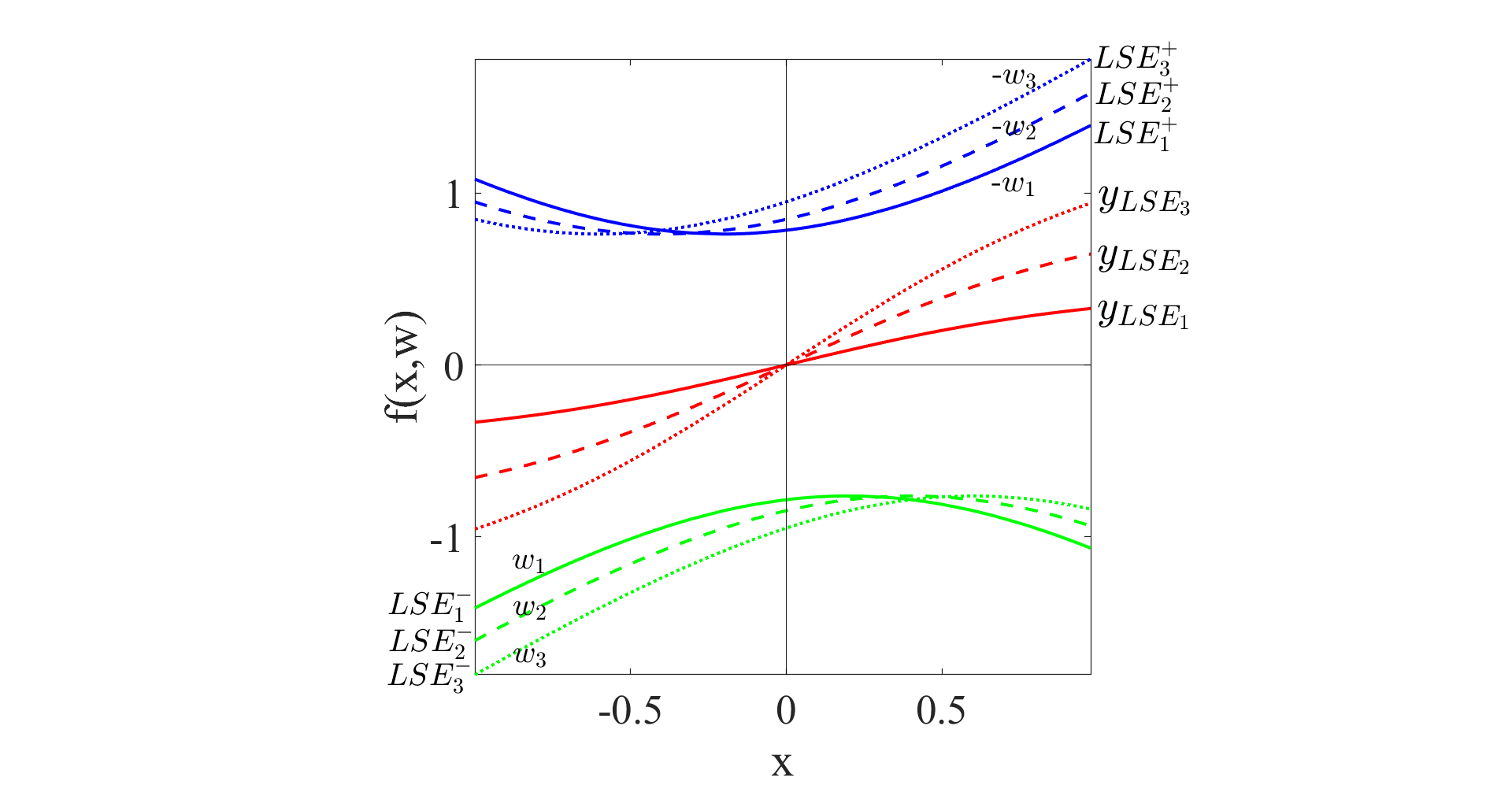}}}
    \subfloat[]{\label{fig:logsum}\stackinset{l}{.05in}{t}{.1in}{(c)}{\includegraphics[page=1,scale=0.208,trim=330 0 300 0,clip]{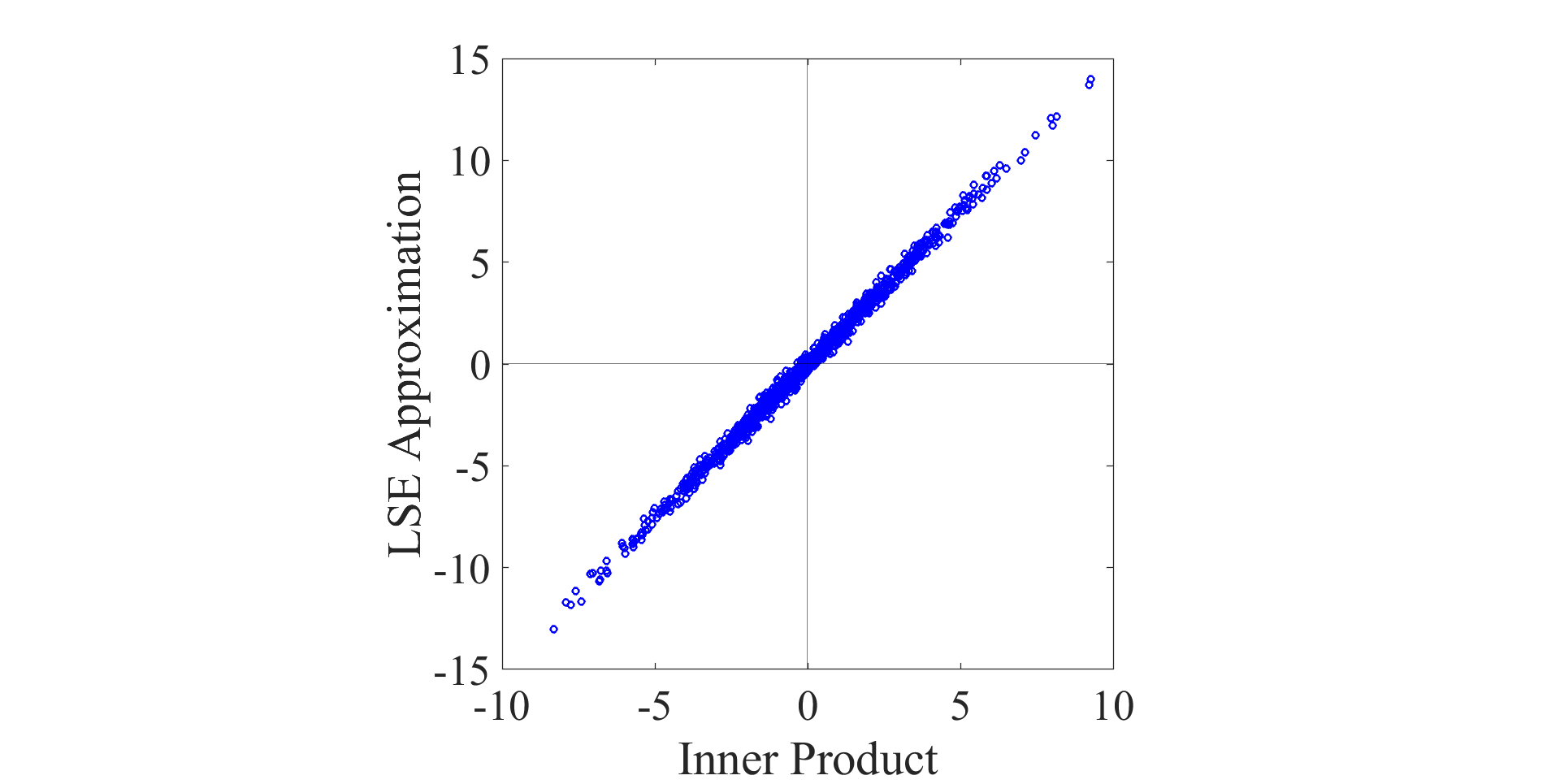}}}
    \caption{(a) Scalar Inner Product expressed in quadratic equation for 3 different $w$, i.e.,($w_1,w_2, w_3$). Here, $QP^+ = (w+x)^2$ and $QP^- = -(w-x)^2$. (b) Scalar Inner Product approximation expressed using log-sum-exponential terms for 3 different $w$, i.e.,($w_1,w_2, w_3$). Here, $LSE^+ = \log \left( e^{w+x} + e^{-w-x} \right)$ and $LSE^- = - \log \left( e^{w-x} + e^{-w+x}\right)$. (c) Inner Product and Log-Sum-Exponential Approximation scatter plot of 64 dimensional vectors with each input randomly varied between -1 and +1 }
\end{figure*}

There have been various neural network implementations that eliminate the usage of multipliers in their algorithms. In \cite{chen2020addernet}, Convolutional Neural Networks (CNNs) are optimized using a new similarity measure by taking $L_1$ norm distance between the filters, and the input features as the output response. Even though it eliminated the multipliers, this implementation requires batch normalization after each convolution operation, resulting in usage of multipliers for this operation. It also requires a full precision gradient-descent training scheme to achieve reasonable accuracy.


Similarly, in \cite{elhoushi2019deepshift}, convolutional shifts and fully connected shifts are introduced to replace multiplications with bitwise shifts and sign flipping. The authors use powers of 2 to represent weights, bias, and activations and use compression logic for storing these values. There is a significant overhead for compression and decompression logic for the implementation of an online system. In \cite{you2020shiftaddnet}, the authors leverage the idea of using additions and logical bit-shifts instead of multiplications to explicitly parameterize deep networks that involve only bit-shift and additive weight layers. This network has limitations for implementing activation functions in the shift-add technique. In all these neural network systems, the training algorithm and activation implementations involve multipliers. Hence these systems cannot be termed as a complete multiplierless system.

\begin{figure*}[ht]
    \centering
    \subfloat[]{\label{fig:1MPwx}\stackinset{l}{.1in}{t}{.1in}{(a)}{\includegraphics[page=1,scale=0.2,trim=330 0 280 0,clip]{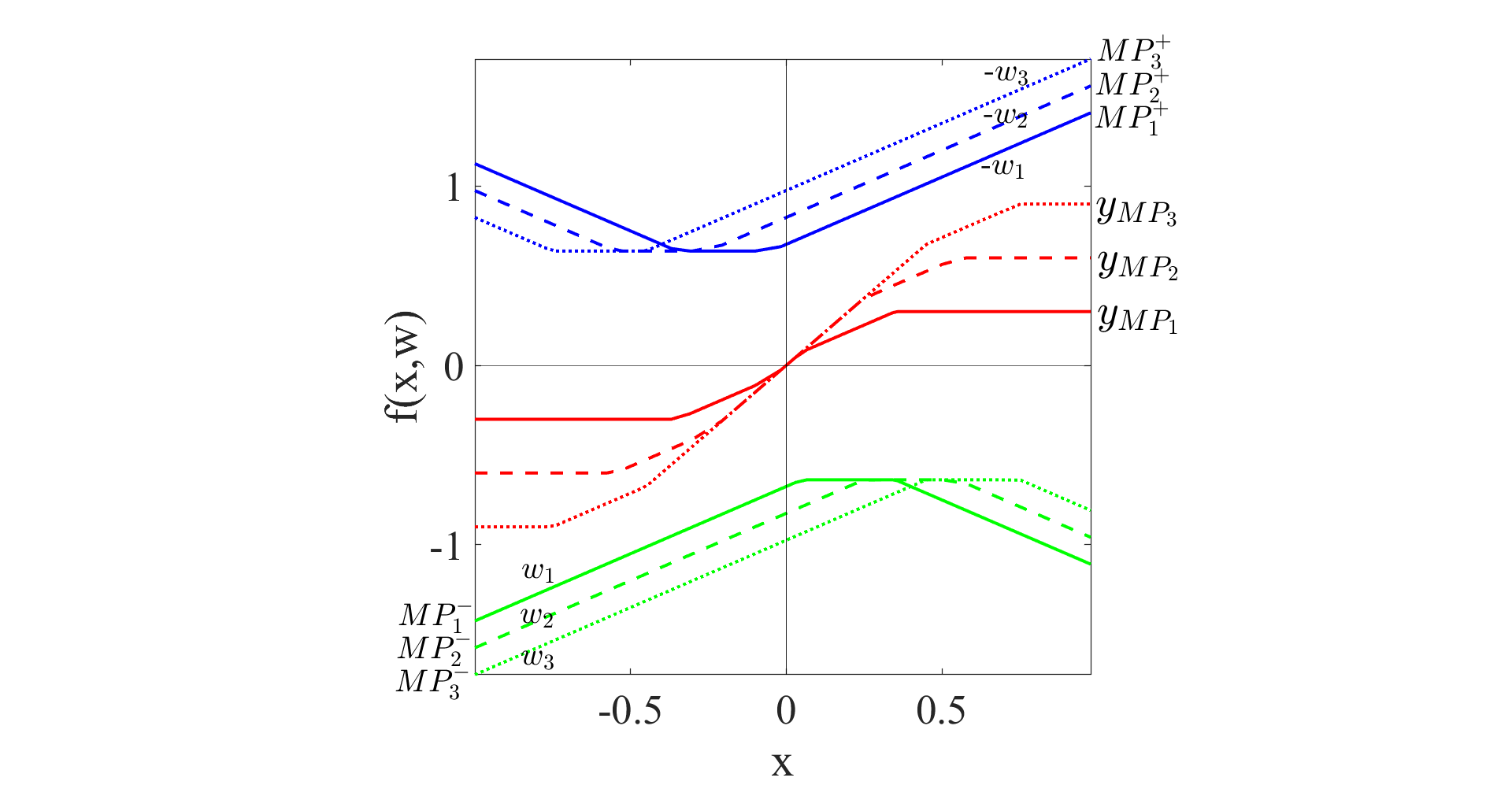}}}
    \subfloat[]{\label{fig:MP}\stackinset{l}{.05in}{t}{.1in}{(b)}{\includegraphics[page=1,scale=0.208,trim=330 0 300 0,clip]{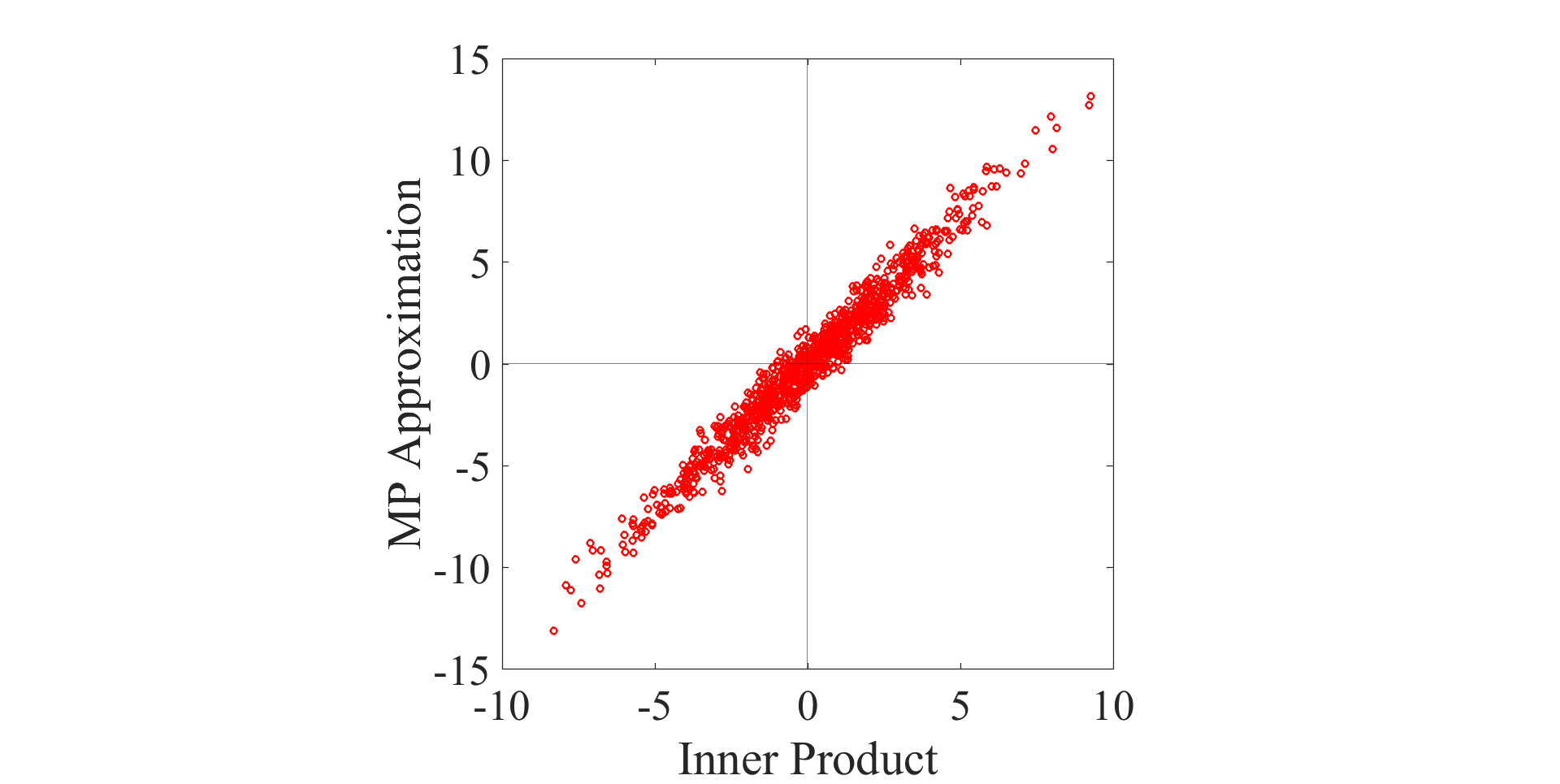}}}
    \subfloat[]{\label{fig:NR}\stackinset{l}{.1in}{t}{.1in}{(c)}{\includegraphics[page=1,scale=0.208,trim=330 0 300 0,clip]{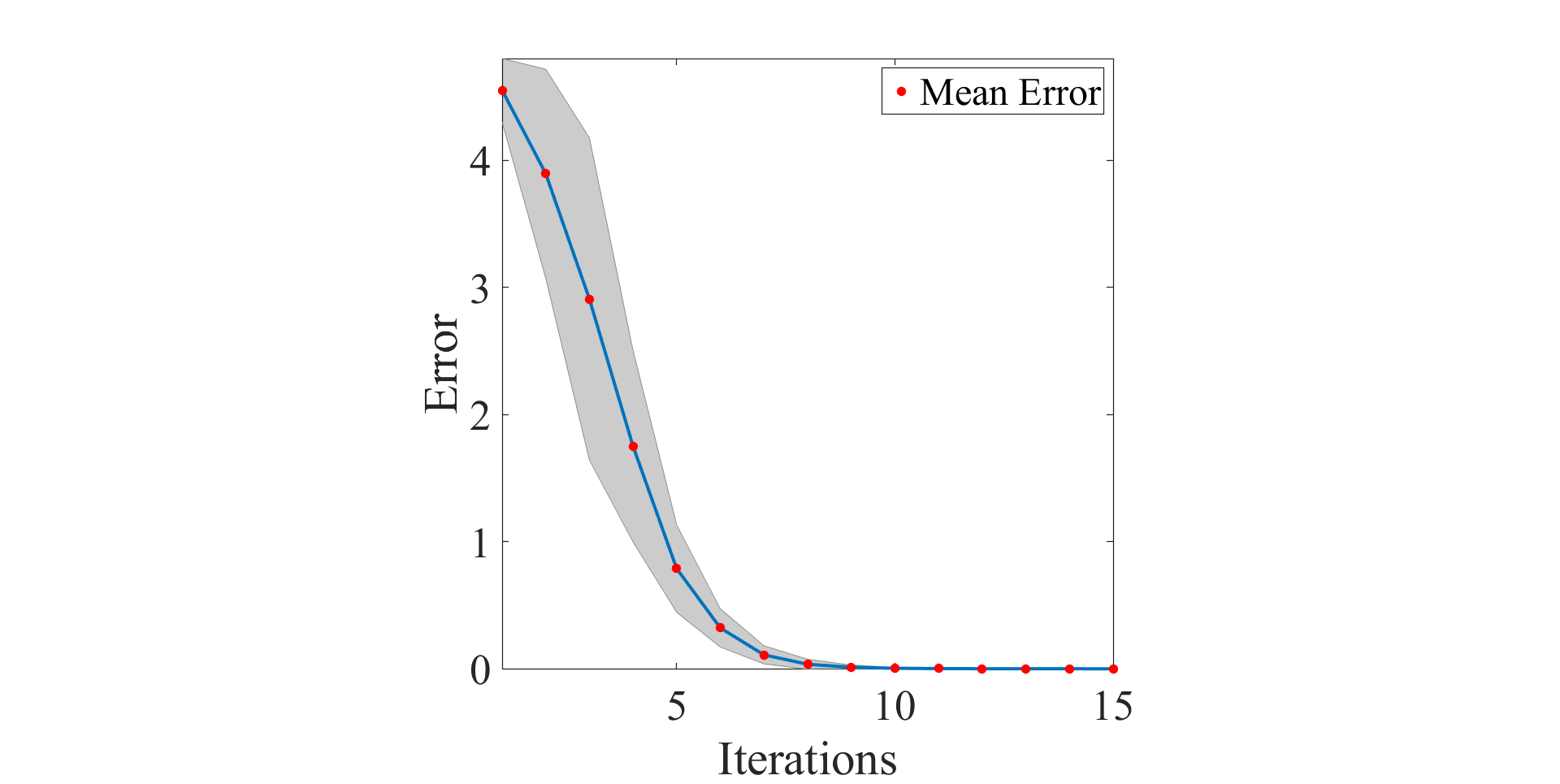}}}
    \caption{(a) Scalar Inner Product approximation expressed in MP domain for 3 different $w$, i.e.,($w_1,w_2, w_3$). Here, $MP^+ = MP\left( \left[ w+x,-w-x \right],\gamma \right)$ and $MP^- = - MP\left( \left[ w-x,-w+x \right], \gamma \right)$. (b) Inner Product and MP Approximation scatter plot of 64 dimensional vectors with each input randomly varied between -1 and +1 (c) Variance in error of $z$ value due to shift approximation of $|S|$ becomes zero after 10 iterations}
\end{figure*}

In this work, we propose to use an MP approximation to implement multiplierless kernel machine. MP-based approximation was first reported in~\cite{chakrabartty2004margin}, and in~\cite{kucher2007energy}, an MP-based SVM was reported for analog hardware. The main objective of this work is to build scalable digital hardware using an optimized MP approximation. Also,
the previous work in MP-based SVM has used offline training. Our system is a one-of-a-kind digital hardware system using a multiplierless approximation technique in conjunction with online training and inference on the same platform.

\section{Multiplierless Inner-product Computation} \label{Multiplierless}
In this section, we first describe a framework to approximate inner-products using a quadratic function, which is then
generalized to the proposed MP-based multiplierless architecture. Consider the following mathematical expression 
\begin{align}
y = \frac{1}{2}\Bigl[f(\w+\x, -\w-\x) \nonumber \\
-f(\w-\x,-\w+\x)\Bigr]. \label{eq:42}
\end{align}
where $f:\mathbb{R}^D \times \mathbb{R}^D \rightarrow \mathbb{R}$ is a Lipschitz continuous function, $y \in \mathbb{R}$ is a scalar variable, $\w \in \mathbb{R}^D$ and $\x \in \mathbb{R}^D$ 
are $D$ dimensional real vectors. 


\textbf{Corollary 1}: If we choose $f$ to be a quadratic equation as $f(\x,-\x) = \frac{1}{2}\x^T\x + c$, where $ c \in \mathbb{R}$ 
is a constant, we get,
\begin{align}
y = \w^T\x. \label{eq:45}
\end{align}
Eq.\eqref{eq:45} is an exact inner-product between the vectors $\w$ and $\x$. The quadratic function is a means to derive the inner product. Inner-products essentially denote Multiply-Accumulate (MAC) operations in any design. One of the most commonly used operations in ML systems including SVMs is MAC operation. We intend to approximate this operation using our MP algorithm with a kernel machine implementation described in Section IV. We can also use this approximation in other systems which utilizes MAC operation like neural networks. The novelty of our proposed work lies in approximating MAC operation to achieve energy-efficient ML systems with minimal accuracy degradation that can be deployed on hardware-constrained devices. Fig.\ref{fig:1Dwx} illustrates this process using two 
scalars $w \in \mathbb{R}$ and $x \in \mathbb{R}$ using a one dimensional quadratic function. 
\begin{align}
y_{QP}(x) = \frac{1}{4}[(w + x)^2 - (w -x)^2]. \label{eq:66}
\end{align}
However, implementing multiplication and inner-products using this approach on digital hardware would require computing a quadratic function, which would require using a look-up table or other numerical techniques \cite{sadeghian2016optimized}.
Also, this approach does not consider the finite dynamic range if the operands are represented using
fixed precision. While the effect of finite precision might not be evident for 16-bit operands, when the precision is
reduced down to 8-bits or lower, $y_{QP}(x)$ will saturate due to overflow or underflow. Next, we consider a form
of $y_{QP}(x)$ that captures the effect of saturation.  

\textbf{Corollary 2}: Let $f$ be a log-sum-exponential (LSE) function defined over the elements of $\x = [x_1,x_2,...,x_D]$ as $f(\x,-\x) = \log \left(\sum \limits_{i=1}^D e^{x_i} + \sum \limits_{i=1}^D e^{-x_i} \right)$, using Taylor-Series  expansion \cite{hlawitschka1994empirical}, we get,
\begin{align}
    f(\w+\x, -\w-\x) \approx f(\x,-\x) + \w^T \nabla f(\x,-\x) \label{eq:42a} \\
    f(\w-\x, -\w+\x) \approx f(\x,-\x) - \w^T \nabla f(\x,-\x) \label{eq:42b}
\end{align}
Substituting these in eq.\eqref{eq:42},
\begin{align}
    y = \sum\limits_{k=1}^D w_k \frac{\sum\limits_{i=1}^D \left[e^{x_i} - e^{-x_i}\right]}{\sum\limits_{i=1}^D \left[e^{x_i} + e^{-x_i}\right]} \approx \w^T\x \label{eq:47} 
\end{align}
The effect of eq.~\eqref{eq:47} can be visualized in Fig.\ref{fig:1Dlogewx} for scalars $w \in \mathbb{R}$ and $x \in \mathbb{R}$ and
for a one-dimensional LSE function
\begin{align}
y_{LSE}(x) = \log \left( e^{w+x} + e^{-w-x} \right) - \log \left( e^{w-x} + e^{-w+x}\right). \label{eq:67}
\end{align}
From Fig.\ref{fig:1Dlogewx}, we see that for smaller values of the operands, the $y_{LSE}(x)$ approximates the 
multiplication operation, whereas for larger values $y_{LSE}(x)$ saturates. This effect also applies to general 
inner-products using multi-dimensional vectors, as described by eq.\eqref{eq:47}.
Fig.\ref{fig:logsum} shows the scatter plot that compares the values of $y$ computed using eq.\eqref{eq:45} and its log-sum-exponential approximation
given by eq.\eqref{eq:47}, for randomly generated $D=64$ dimensional vectors $\w$ and $\x$. The plot clearly shows that the two functions approximate
each other, particularly for a smaller magnitude of $||\x||$. Like the quadratic function, implementing the LSE function on digital hardware would also require look-up tables. Note that other choices of 
$f(.)$ could also lead to similar multiplierless approximations of the 
inner-products. However, we are interested in finding the 
$f$ that can be easily implemented using simple digital hardware primitives.

\begin{figure*}[ht]
    \centering
    \subfloat[]{\label{fig:16-bit}\stackinset{l}{.03in}{t}{.08in}{(a)}{\includegraphics[page=1,scale=0.18,trim=350 0 360 0,clip]{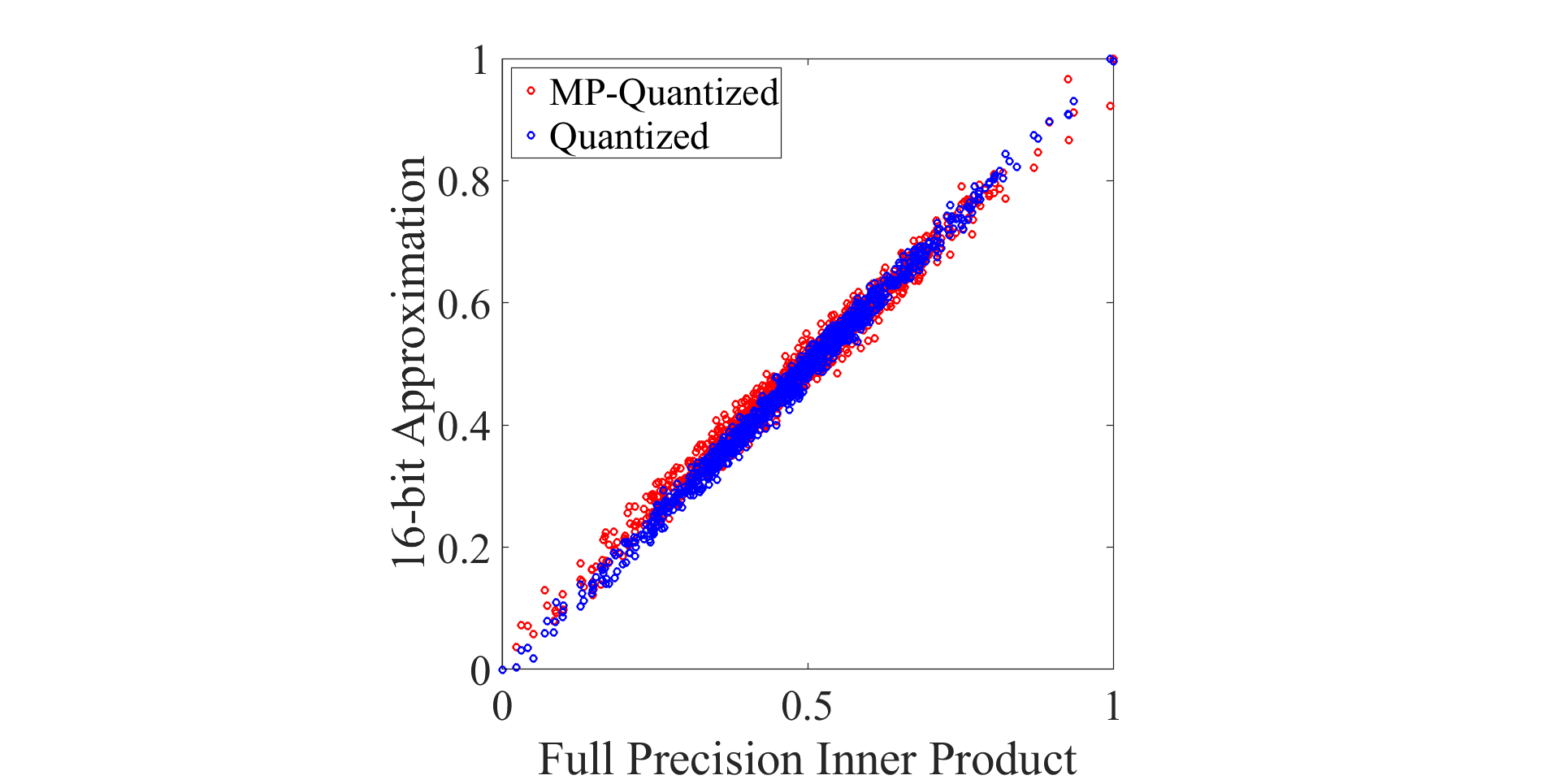}}}
    \subfloat[]{\label{fig:14-bit}\stackinset{l}{.03in}{t}{.08in}{(b)}{\includegraphics[page=1,scale=0.18,trim=350 0 360 0,clip]{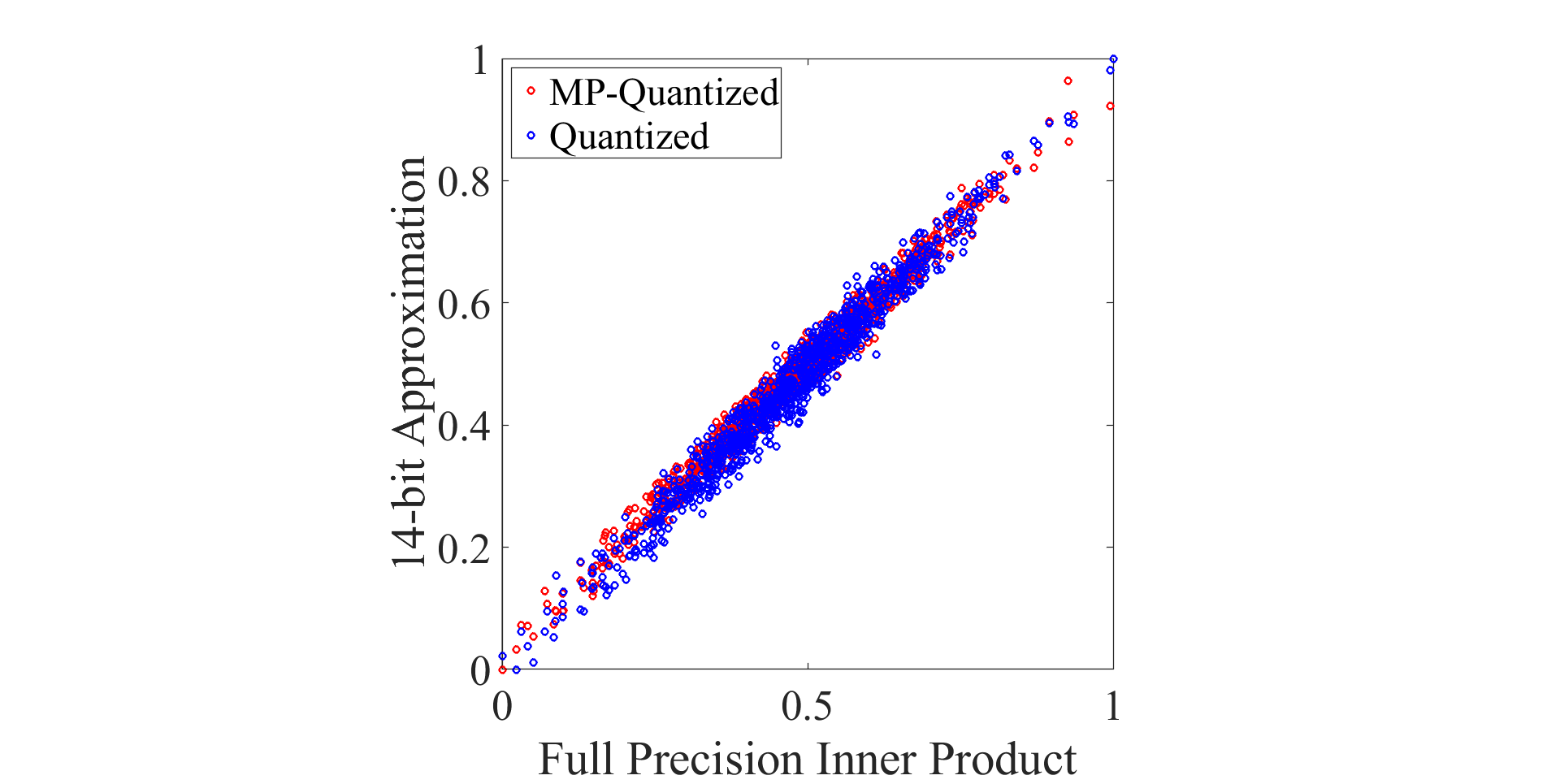}}}
    \subfloat[]{\label{fig:12-bit}\stackinset{l}{.03in}{t}{.08in}{(c)}{\includegraphics[page=1,scale=0.18,trim=350 0 360 0,clip]{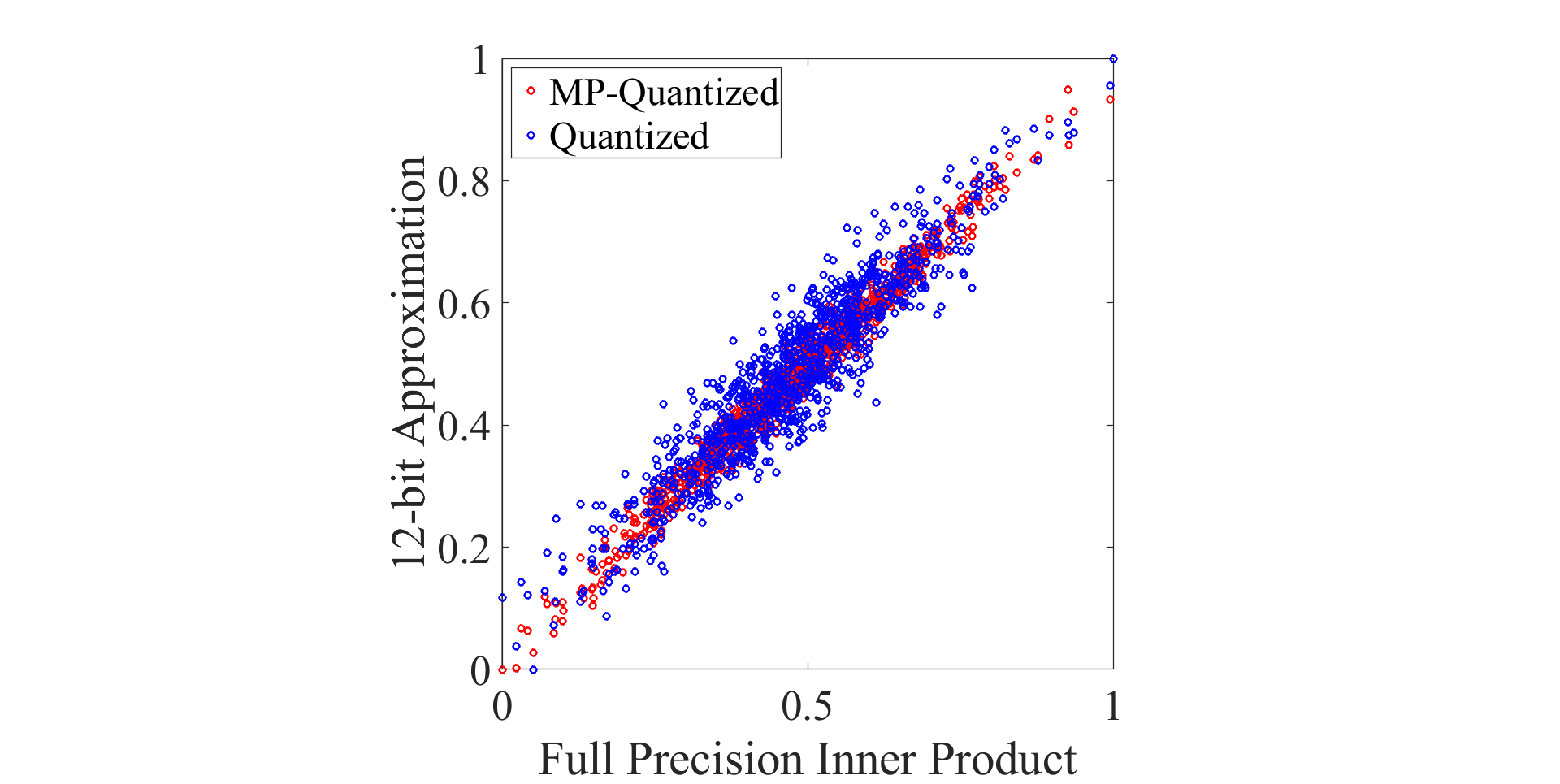}}}
    \subfloat[]{\label{fig:10-bit}\stackinset{l}{.03in}{t}{.08in}{(d)}{\includegraphics[page=1,scale=0.18,trim=350 0 360 0,clip]{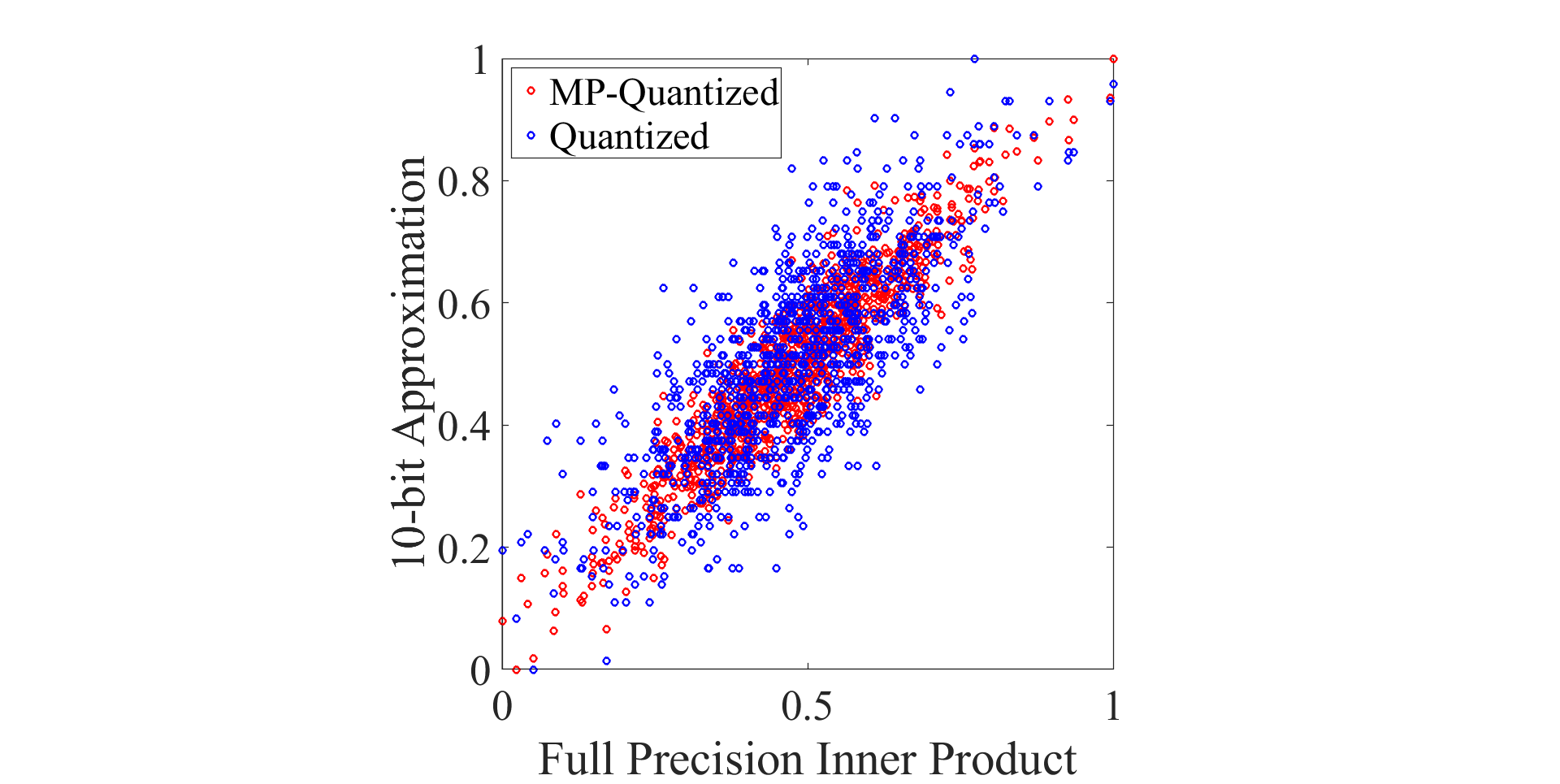}}}
    \caption{Impact of quantization and approximation on inner product normalized between 0 and 1 for 1000 pairs of 64 dimensional vector randomly sampled between -1 and 1.}
    \label{fig:Quantized}
\end{figure*}

\subsection{Margin Propagation based Multiplierless Approximation} \label{MP_Approx}
Margin Propagation is an approximating technique which can be used to generate different forms of linear and nonlinear functions using thresholding operations \cite{gu2012theory}. Here, we explain the steps to derive this technique.
We first express the LSE equation from Corollary 2 as, 
\begin{align}
z_{log} = \gamma \log\left(\sum \limits_{i=1}^D e^{\frac{x_i}{\gamma}} + e^{\frac{-x_i}{\gamma}}\right). \label{eq:48}
\end{align}
as a constraint 
\begin{align}     
\sum \limits_{i=1}^D e^{\frac{[x_i-z_{log}]}{\gamma}} + \sum \limits_{i=1}^D e^{\frac{[-x_i-z_{log}]}{\gamma}} = 1. \label{eq:49} 
\end{align}
Here, $\gamma > 0$ is a hyper-parameter.
In the first-order MP-approximation, the exponential function is approximated using a simple piecewise linear model as
\begin{align}
e^{\frac{x-z_{log}}{\gamma}} \approx \bigl[\frac{x-z_{MP}}{\gamma} +1\bigr]_+. \label{eq:50} 
\end{align}
where $[\cdot]_+ = max(\cdot,0)$ is a rectifying linear operation, $z_{MP}$ is the approximation of $z_{log}$. The detailed explanation of piecewise linear approximation of log-sum-exponential function is described in \cite{gu2012theory}.
Thus, the constraint in eq.\eqref{eq:49}
can be written as 
\begin{align}
\sum \limits_{i=1}^D\left[x_i-z_{MP}+\gamma\right]_+ + \sum \limits_{i=1}^D\left[-x_i-z_{MP}+\gamma\right]_+ = \gamma. \label{eq:51}
\end{align}
Here, $z_{MP}$ is computed as the solution of the equation and $z_{MP} \cong MP(\x,\gamma)$ forms the MP function.
Note that all operations in eq.\eqref{eq:51} requires only
unsigned arithmetic, and $[\cdot]_+$ operation. $[\cdot]_+$ operation is essentially a ReLU operation which has used in digital hardware by implementing a register underflow. So MP-function could be readily and efficiently mapped onto low-precision digital hardware for computing inner-products and multiplications. The detailed derivation of the MP-function is described as a reverse water-filling procedure in \cite{gu2012theory}. 
Based on eq.\eqref{eq:42}, eq. \eqref{eq:45} and \eqref{eq:51}, we get 
\begin{align}
    y = \w^T\x \approx MP([\w+\x, -\w-\x],\gamma) \quad \nonumber \\ 
    - MP([\w-\x, -\w+\x],\gamma). \label{eq:89} 
\end{align}
The approximation of a multiplication operation can be visualized for scalars $w \in \mathbb{R}$ and $x \in \mathbb{R}$, for a one-dimensional MP-function as
\begin{align}
y_{MP}(x) = MP\left( \left[ w+x,-w-x \right],\gamma \right) \quad \nonumber \\ 
- MP\left( \left[ w-x,-w+x \right], \gamma \right). \label{eq:68}
\end{align}
Fig.\ref{fig:1MPwx} shows that the MP-function computes a piecewise linear approximation of the LSE function and exhibits the saturation due to register overflow/underflow.  

When the function $f(.)$ in eq.\eqref{eq:42} is replaced by $MP(\x,\gamma)$, an approximation to the inner-product can be obtained using only piecewise linear functions. Fig.\ref{fig:MP} shows the scatter plot that compares the true inner-product with the MP-based inner-product, showing that the approximation error is similar to that of the log-sum-exp approximation in Fig.\ref{fig:logsum}. Thus, MP-function serves as a good approximation to the inner-product computation.

\subsection{Implementation of MP-function on Digital Hardware}

Given an input vector $\x = \{x_1, x_2, ..., x_D\}$ and hyper-parameter $\gamma$, in~\cite{gu2012theory} an algorithm was presented to compute the function $z = MP(\x,\gamma)$. The approach was based on iterative identification of the set $S = \{x_i; x_i > z \}$ using the reverse water filling approach. From eq.\eqref{eq:51}, we get the expression
of $MP(\x,\gamma)$ as
\begin{equation}
    MP(\x,\gamma) = - \frac{\gamma}{|S|} + 
        \frac{ \sum\limits_{i \in S} x_i }{|S|}. \label{eq:55}
\end{equation}
where $|S|$ is the size of the set $S$. Since $|S|$ is a function of $MP(\x,\gamma)$,
the expression in eq.\eqref{eq:55} requires dividers.

\begin{figure}[!ht]
\centerline{\includegraphics[page=1,scale=0.6,trim=0 0 0 0,clip]{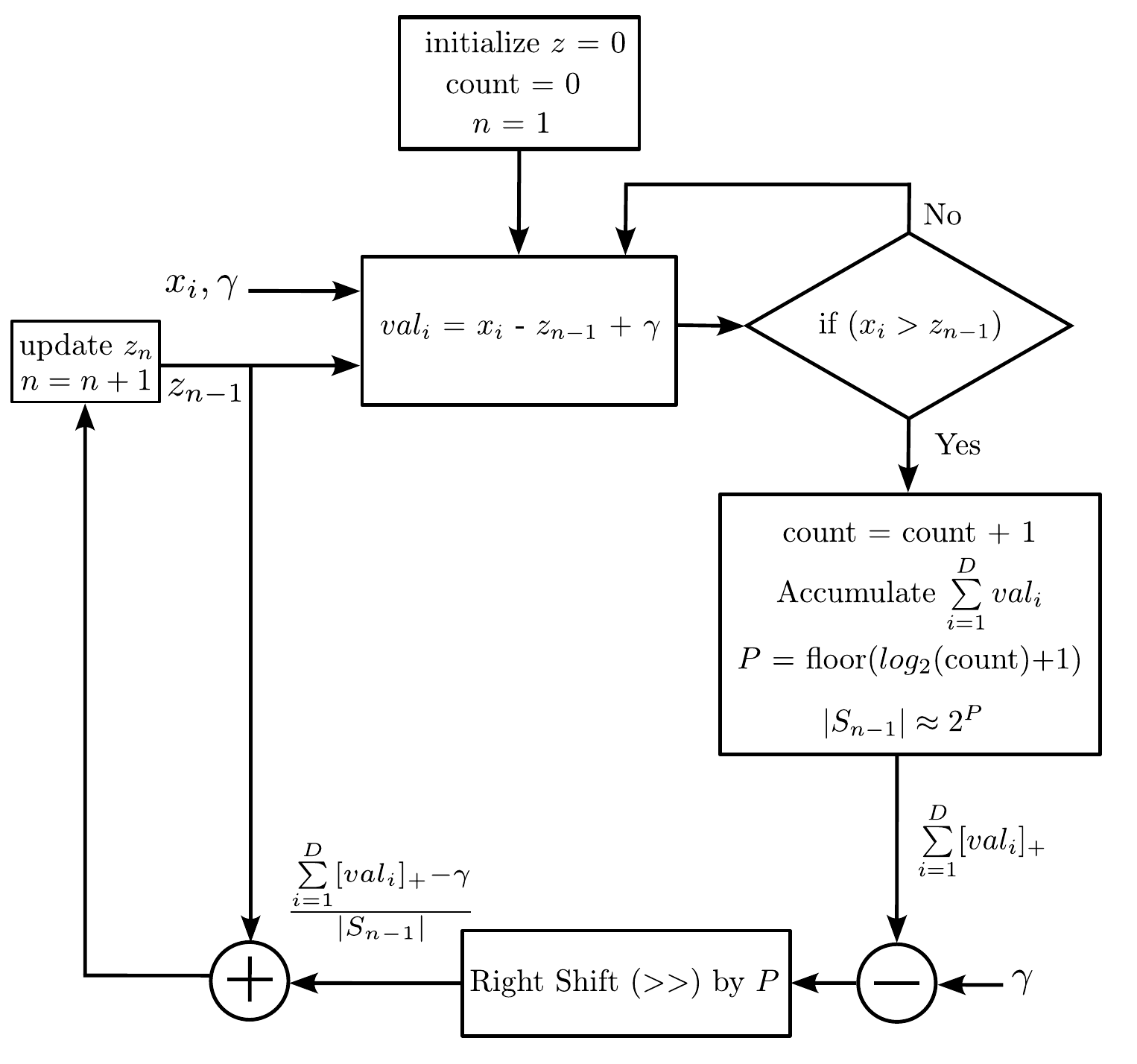}}
\caption{Newton Raphson method based MP formulation. ${|S|}$ is the approximated value of variable \textit{count} and represented as the nearest upper bound power of 2.}
\label{algo:NR}
\end{figure}

We now report an implementation of MP-function that uses only basic digital hardware
primitives like addition, subtraction, shift, and comparison operations. The implementation poses the constraint in eq.\eqref{eq:51} as a root-finding problem for the variable $z$. Then, applying Newton-Raphson recursion \cite{ben1966newton}, the MP function can be iteratively computed as 
\begin{equation}
z_n \leftarrow z_{n-1} + \frac{\sum\limits_{i=1}^D\left[x_i-z_{n-1} + \gamma \right]_+ - \gamma}{|S_{n-1}|}. \label{eq:56}
\end{equation}
Here, $S_n = \{x_i; x_i > z_n \}$ and $z_n \xrightarrow{{n \rightarrow \infty}} MP(\x,\gamma)$. The Newton-Raphson step can be made hardware-friendly by quantizing $S_{n-1}$ to the nearest upper bound of power of two as $S_{n-1} \approx 2^{P_{n-1}}$ or by choosing a fixed power of 2 in terms of $P$. Here, $P = floor(log_2(count)) + 1$ as shown in Fig~\ref{algo:NR}. This is implemented in hardware using a priority encoder that checks the highest bit location whose bit value is $1$ for the variable $count$. 
Then, the division operation in eq.\eqref{eq:56} can be replaced by a right-shift operation. This approximation error can be readily tolerated within each Newton-Raphson step since each step
will correct any approximation error in the previous step. Note that the Newton-Raphson
iteration converges quickly to the desired solution, and hence only a limited number of iterations are required in practice.
Fig.\ref{fig:NR} shows several examples of the MP-function converging to the final value within 10 Newton-Raphson iterations for a
100-dimensional input $\x$. Thus, in our proposed algorithm in Fig~\ref{algo:NR} used for computing the MP-function we limit the number of
Newton-Raphson iterations to 10. 

\begin{table}[]
\centering
\caption{Bit width comparison for both types of quantization used in Fig.~\ref{fig:Quantized}. We use higher precision MP inputs for the same bit width inner product output, resulting in better outputs.\label{InnerProductCompare}}
\begin{tabular}{|c|c|c|}
\hline
\multirow{2}{*}{\textbf{\begin{tabular}[c]{@{}c@{}}Inner product bit width \\  (64-dimensions)\end{tabular}}} & \multicolumn{2}{c|}{\textbf{Input bit width}} \\ \cline{2-3} 
                                                                                             & \textbf{Quantized}   & \textbf{MP-Quantized}  \\ \hline \hline
16-bit                                                                                       & 5-bit                & 9-bit                  \\ \hline
14-bit                                                                                       & 4-bit                & 7-bit                  \\ \hline
12-bit                                                                                       & 3-bit                & 5-bit                  \\ \hline
10-bit                                                                                       & 2-bit                & 3-bit                  \\ \hline
\end{tabular}
\end{table}

In order to make any algorithm hardware-friendly, the classic approach is to express it in minimum possible bit width precision with minimal loss in functionality. This would help reduce area and power when implementing digital hardware. In Fig.~\ref{fig:Quantized}, we see the impact of reducing the bit precision on the approximation of inner product for a 64-dimensional input vector sampled between -1 and 1. We see the variance in inner product for MP approximation exists, which can be termed as the MP approximation error, compared to the standard quantization of the inner product even for higher bit precision (Fig.~\ref{fig:Quantized}(a) and ~\ref{fig:Quantized}(b)). We use the online learning approach, detailed in Section \ref{On-chip}, for MP approximation in our system with minimal hardware increase to mitigate this approximation error. However, as we reduce the bit precision further, we see the quantization error increases and overlaps the MP approximation error (Fig.~\ref{fig:Quantized}(c) and ~\ref{fig:Quantized}(d)). We see in Table~\ref{InnerProductCompare}, for an $n$-bit inner product output, the standard quantized version requires $\frac{(n-6)}{2}$~-~bit input vector, whereas quantized MP inner product requires $(n-6-1)$~-~bit input vector due to addition operation instead of multiplication. Here, we use 6-bits for the 64-dimensional input vector. This shows that MP has better input precision for the same output bit width, leading to better output approximation. The online learning in our design helps mitigate the MP approximation errors by adjusting the weights to accommodate these errors. Thus, the MP approximation, despite quantization, performs better or is equivalent to the standard quantized approach as we reduce the bit precision of the output. At the same time, MP uses lesser hardware for the output with the same bit width inputs as it uses simple hardware elements like adders in comparison to multipliers for the standard approach.  
Hence, MP proves to be a hardware-friendly approximation even when the algorithm is quantized. 






\begin{figure*}[ht]
\centerline{\includegraphics[page=1,scale=0.31,trim=0 0 0 0,clip]{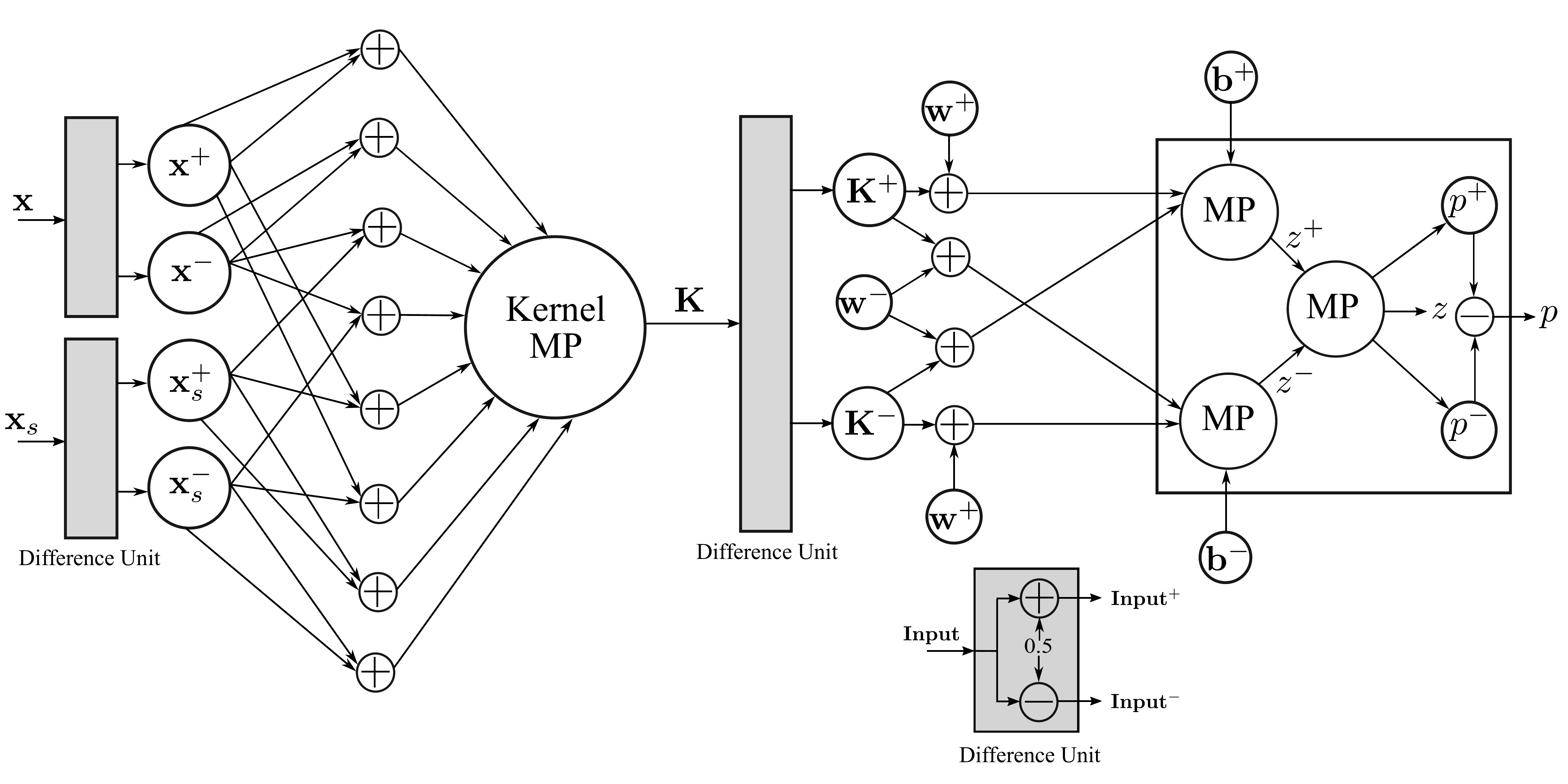}}
\caption{ MP kernel machine architecture. Kernel MP is based on eq. \eqref{eq:15} and the other MP functions are described from eq. \eqref{eq:38} to \eqref{eq:40}.}
\label{Top_Level_Math}
\end{figure*} 

\subsection{Energy-cost of MP-based Computing}

Let $C, C_M$, and $C_A$ denote the total energy-cost corresponding to an MVM operation, a single multiplication operation, and a single addition operation, respectively. Then, MVM of a $1 \times M$ vector and a $M \times M$ matrix would incur an energy-cost  
\begin{align}
    C = M^2 \times C_M + (M^2 - M) \times C_A. \label{eq:61}
\end{align}
For an MP-based approximation of the MVM, the energy cost incurred is
\begin{align}
    C_{MP} =  (M^2 + (M \times F \times R)) \times C_A +  M \times R \times C_C. \label{eq:63}
\end{align}
Here, $F$ is the sparsity factor determined by $\gamma$, typically having a value less than 1, and $C_C$ is the energy-cost for a comparison operation having $\mathcal{O}(1)$ complexity. Also, note that $R$ is the number of Newton-Raphson iterations, which is $10$ in our case. Thus, as inputs increase beyond $R$, MP approximation complexity reduces further compared to that of MVM.
Multiplication requires $n^2$ full adders for $n$-bit system, i.e., $C_M = n^2 \times C_A$ \cite{vittoz1990future}. $C_A$ has linear complexity, i.e., $\mathcal{O}(n)$. Hence, the MP approximation technique becomes ideal for digital systems as the implementation of adder logic is less complex and low on resource usage than the equivalent multiplier.

\section{MP Kernel Machine Inference} \label{MP_Math}

We now use the MP-based inner product approximation to design an MP kernel machine. Consider a vector ${\x} \in \mathbb{R}^d$, the decision function for kernel machines~\cite{cristianini2000introduction} is given as,
\begin{align}
    f({\x}) = {\w}^T{\K} + \mathbf{b}. \qquad \qquad \label{eq:5}
\end{align}
where $f:\mathbb{R}^d \rightarrow \mathbb{R}$,  $ \K :\mathbb{R}^{d \times d} \rightarrow \mathbb{R}^d$ is the kernel which is a function of $\x$, and  $\w \in \mathbb{R}^d$, $\mathbf{b} \in\mathbb{R}$ is the corresponding trained weight vector and bias, respectively. 
As, MP approximation, as shown in eq.~\eqref{eq:68}, is in differential format, we express the variables as $\w = \w^+ - \w^-$, $\mathbf{b} = \mathbf{b^+} - \mathbf{b^-}$ and $\K = \K^+ - \K^-$. 
\begin{align}
    f({\x}) = {(\w^+ - \w^-)}^T(\K^+ - \K^-) + (\mathbf{b^+} - \mathbf{b^-}). \qquad \nonumber \\
    f({\x}) = \left[{(\w^+)}^T\K^+ + {(\w^-)}^T\K^- + \mathbf{b^+}\right] \; \qquad \quad \qquad \nonumber \\
            - \left[{(\w^+)}^T\K^- + {(\w^-)}^T\K^+ + \mathbf{b^-}\right]. \: \: \quad \qquad \quad \label{eq:69}
\end{align}
Using eq.\eqref{eq:45}, and applying MP approximation based on eq.\eqref{eq:68}, we can express eq.\eqref{eq:69} as,
\begin{align}
    f_{MP}(\x) = MP([\w^{+} + \K^{+}, \w^{-} + \K^{-}, \mathbf{b^+} ],\gamma) \nonumber \\
    -MP([\w^{+} + \K^{-} , \w^{-} + \K^{+} ,\mathbf{b^-} ],\gamma).  \label{eq:9}
\end{align}
Fig.\ref{Top_Level_Math} describes the kernel machine architecture using MP approximation. The input is provided to a difference unit to generate $\x^+,\x^-,\x_s^+$ and $\x_s^-$ vectors. The kernel MP generates the kernel output with inputs as a combination of $\x^+,\x^-,\x_s^+,\x_s^-$ based on the kernel used. In our case, we use the kernel mentioned in next section \ref{kfunc}.
This kernel $\K$ is used to produce $\K^+$ and $\K^-$ with the help of the difference unit.
The weights and bias generated, described in section \ref{On-chip}, are used with the kernel combination to generate MP approximation output as below.
We can express eq.\eqref{eq:9} as,
\begin{align}
    f_{MP}(\x) = z^+ - z^-. \label{eq:41}
\end{align}
where,
\begin{align}
    z^+ = MP([\w^+ + \K^+, \w^- + \K^-, \mathbf{b^+} ],\gamma_1). \label{eq:38} \\
    z^- = MP([\w^+ + \K^-, \w^- + \K^+, \mathbf{b^-} ],\gamma_1). \label{eq:39}
\end{align}
$\gamma_1$ is a hyper-parameter that is learned using gamma annealing described in Algorithm \ref{algo:gamma}.
We normalize the values for $z^+$ and $z^-$ for better stability of the system using MP,
\begin {align}
    z = MP([z^+,z^-],\gamma_n). \label{eq:40}
\end{align}
Here, $\gamma_n$ is the hyper-parameter used for normalization. In this case, $\gamma_n = 1$.
The output of the system can be expressed in differential form,
\begin{align}
    p = p^{+} - p^{-}. \label{eq:16}
\end{align}
Here, $p \in \mathbb{R}$, $p^{+} + p^{-} = 1$ and $p^+,p^- \geq 0$.
As $z$ is the normalizing factor for $z^+$ and $z^-$, we can estimate the output using reverse water filling algorithm \cite{gu2012theory}, which is generated by the MP function in eq.\eqref{eq:40} for each class,
\begin{align}
    p^{+} = [z^{+} - z]_{+}. \nonumber \\
    p^{-} = [z^{-} - z]_{+}. \label{eq:17}
\end{align}

\subsection{MP Kernel Function} \label{kfunc}
We use a similarity measure function, which has similar property as a Gaussian function, between the vectors $\x$ and $\x_s$ for defining Kernel MP approximation as,

\begin{align}
    \K^- = MP \: ( [2\x_{s}^{+} , 2\x_{s}^{-}, 2\x^{+}, 2\x^-, \quad \quad \nonumber \\
                \x_{s}^+ + \x^- + 2, \; \; \qquad \qquad \nonumber \\
                \x_{s}^- + \x^+ + 2 ], \gamma_2 ). \; \; \quad \quad \label{eq:15}
\end{align}
$\gamma_2$ is the MP hyper-parameter for kernel. We define $\K^+$ = $-\K^-$. The kernel function is derived in detail in Section A of the supplementary document.

Complexity of Kernel Machine, based on eq.\eqref{eq:61} and \eqref{eq:5}, can be expressed as,
\begin{align}
    C_{KM} =  (M^2 + M) \times C_M + (M^2 + M - 1)\times C_A.\label{eq:64}
\end{align}
 and similarly, complexity of kernel machine in MP domain based on eq.~\eqref{eq:63} can be expressed as,
 \begin{align}
    C_{MPKM} = (F \times R \times M + 2 \times  M^2 + 1) \times  C_A \nonumber \\
                + \; R \times M \times C_c. \qquad \qquad \qquad \qquad \label{eq:65}
\end{align}
The complexity equations show that the MP kernel machine complexity is a fraction of that of traditional SVM. This can be leveraged to reduce the power and hardware resources and increase the speed of operation of the MP kernel machine system over traditional SVM.

\section{Online training of MP Kernel Machine} \label{On-chip}


The training of our system requires cost calculation and parameter updates to be done over multiple iterations ($\tau$) using the gradient descent approach, which is described below.

\IEEEPARstart{} Consider a two-class problem, and the cost function can be written as,
\begin{align}
    E = \sum\limits_{n=1}^M |y_{n}^{+} - p_n^{+}| + |y_{n}^{-} - p_n^{-}|. \label{eq:20}
\end{align}
where $M$ is the number of input samples,
$y_n^+ $  and $y_n^-$ are the class labels where $y_n^+, y_n^- \in {0,1}$
$p_n^{+}$ and $p_n^{-}$  are the respective predicted values. $p_n^+,p_n^- \geq 0$.
We have selected the absolute cost function as it is easier to implement on hardware, as we require fewer bits to represent this cost function than the squared error function.

\begin{figure*}[ht]
\centerline{\includegraphics[page=1,scale=0.54,trim=0 0 0 0,clip]{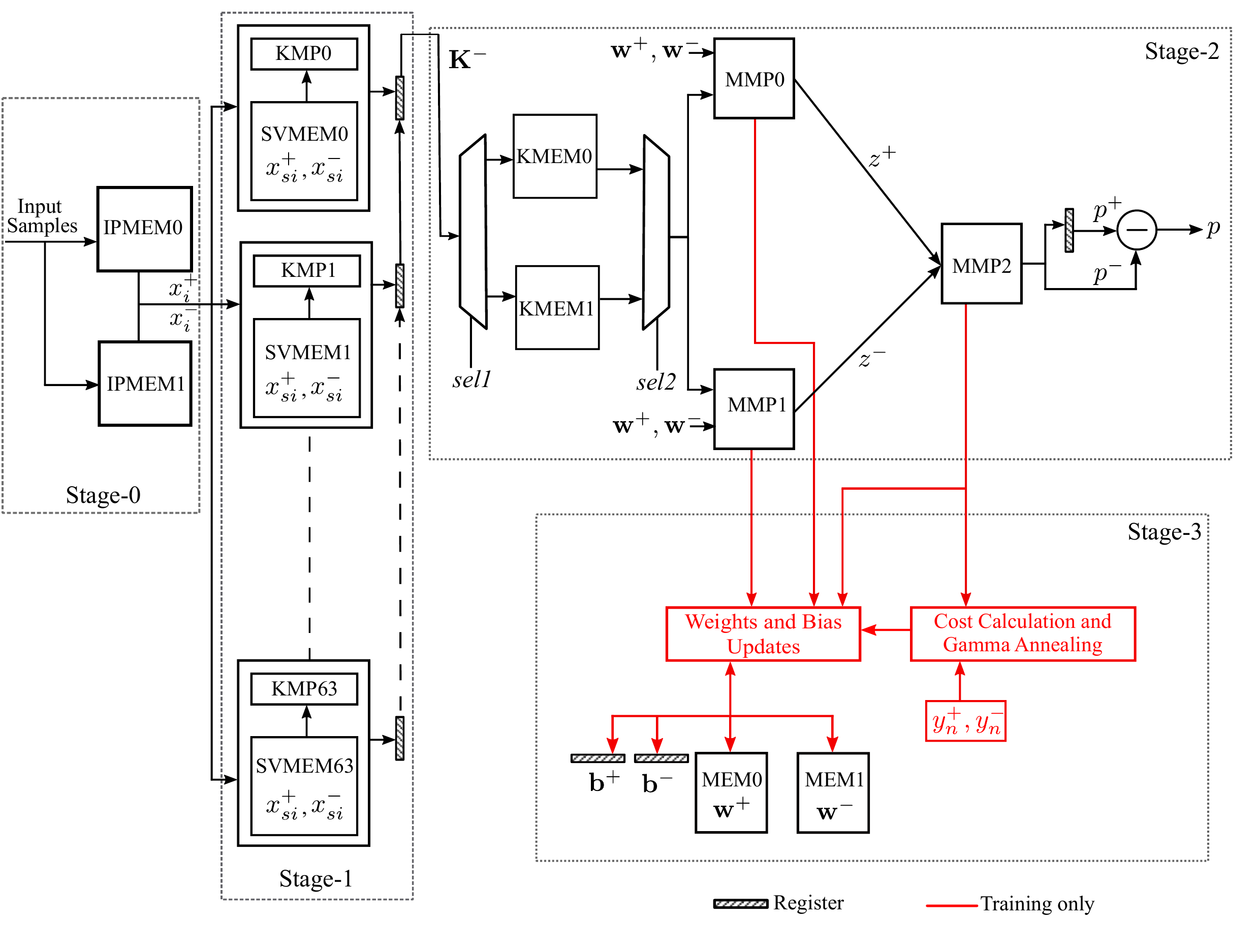}}
\caption{ High-level block diagram of MP kernel machine.  The proposed design shares hardware resources during both the training and inference phase. The red blocks are used only during training.}
\label{top_mpsvm}
\end{figure*}

The weights and biases are updated during each iteration using gradient descent optimization,
\begin{align}
    \w^+ \leftarrow \w^+ - \eta \biggl[ \sum\limits_{n=1}^M \biggl( (\sgn^{+})\left(1 - \frac{1}{|S|}\right)\biggl(I_{(z^+)}\frac{1}{|S_p|}I_{(w^+K^+)}   
     \nonumber \\  
    - \frac{1}{|S_n|}I_{(w^+K^-)}\biggr) \nonumber \\
    + \quad (\sgn^{-})  \left(1 - \frac{1}{|S|}\right) 
    \biggl(I_{(z^-)}\frac{1}{|S_n|}I_{(w^+K^-)}  \nonumber \\
    - \frac{1}{|S_p|}I_{(w^+K^+)}\biggr) \biggr) \biggr]. \label{eq:33} 
\end{align}
\begin{align}
    \w^- \leftarrow \w^- - \eta\biggl[\sum\limits_{n=1}^M  \biggl((\sgn^{+})\left(1 - \frac{1}{|S|}\right) \biggl(I_{(z^+)}\frac{1}{|S_p|}I_{(w^-K^-)} \nonumber \\  
    - \frac{1}{|S_n|}I_{(w^-K^+)}\biggr) \nonumber \\ 
   + \quad  (\sgn^{-}) \left(1 - \frac{1}{|S|}\right) 
   \biggl(I_{(z^-)}\frac{1}{|S_n|}I_{(w^-K^+)} \nonumber \\
   - \frac{1}{|S_p|}I_{(w^-K^-)}\biggr) \biggr)\biggr]. \label{eq:34}
\end{align}
\begin{align}
    \mathbf{b^+} \leftarrow \mathbf{b^+} - \eta\biggl[ \sum\limits_{n=1}^M (\sgn^+) \left(1 - \frac{1}{|S|}\right)I_{(z^+)}\frac{1}{|S_p|}I_{\mathbf{(b^+)}} \biggr] .\label{eq:35}
\end{align}
\begin{align}
    \mathbf{b^-} \leftarrow \mathbf{b^-} - \eta\biggl[\sum\limits_{n=1}^M (\sgn^-) \left(1 - \frac{1}{|S|}\right)I_{(z^-)}\frac{1}{|S_n|}I_{\mathbf{(b^-)}} \biggr]. \label{eq:36}
\end{align}
where $\eta$ is the learning rate, 
$\sgn^{+}$ = $\sgn{(p_n^+ - y_n^+)}$, 
$\sgn^{-}$ = $\sgn{(p_n^- - y_n^-)}$,
$I_{(z^+)}$ = $\mathbbm{1}{(z^+ > z)}$, 
$I_{(z^-)}$ = $\mathbbm{1}{(z^- > z)}$, 
$I_{(w^+K^+)}$=$\mathbbm{1}{(\w^{+} + \K^{+} > z^+)}$,
$I_{(w^+K^-)}$=$\mathbbm{1}{(\w^{+} + \K^{-} > z^-)}$,~
$I_{(w^-K^+)}$=$\mathbbm{1}{(\w^{-} + \K^{+} > z^+)}$,
$I_{(w^-K^-)}$=$\mathbbm{1}{(\w^{-} + \K^{-} > z^-)}$,
$I_{\mathbf{(b^+)}}$ = $\mathbbm{1}{(\mathbf{b^+} > z^+)}$ 
and $I_{\mathbf{(b^-)}}$ = $\mathbbm{1}{(\mathbf{b^-} > z^-)}$. 
$\mathbbm{1}$ is the indicator function. An indicator function is defined on a set $X$ indicating membership of an element in a subset $A$ of $X$, having the value $1$ for all elements of $X$ in $A$ and value $0$ for all elements of $X$ not in $A$.
The gradient descent steps have been derived in detail in Section B of the supplementary document


\begin{algorithm}
\SetAlgoLined
\KwInput{$E_{\tau}$,$\delta$,$\epsilon$,$iter$} 
\KwOutput{$\gamma_1$}
Intialize     $\gamma_1 = c$  \tcp*{Initalize to a value $c$ based on input for MP function}
\For {$\tau = 2$ to $iter$}{ 
\If{($E_{\tau - 1} - E_{\tau}) > \delta$}
{
$\gamma_1 = \gamma_1 - \epsilon$\;   
}
}


\caption{Gamma Annealing. $E_{\tau}$ is the value of cost function \eqref{eq:20} estimated at iteration $\tau$. Here, $\epsilon$ and $\delta$ are empirical values based on the input dataset. $iter$ is the number of iteration for the MP gradient update.}

\label{algo:gamma}
\end{algorithm}

The parameter $\gamma$ in eq.\eqref{eq:55} impacts the output of MP function and hence can be used as a hyper-parameter in the learning algorithm for the 2\textsuperscript{nd} layer as per Fig.\ref{Top_Level_Math}, i.e., $ \gamma_1 $. This value is adjusted each iteration based on the change in the cost between two consecutive iterations as described in Algorithm \ref{algo:gamma}.

Since the gradient is obtained using the MP approximation technique, the training of the system is more tuned to minimizing the error rather than mitigating the approximation itself and hence improves the overall accuracy of the system.



As we see in all these equations, we require only basic primitives such as comparator, shift operators, counters, and adders to implement this training algorithm, making it very hardware-friendly in terms of implementation.

\section{FPGA Implementation of MP Kernel Machine} \label{FPGA}

The FPGA implementation of the proposed multiplierless kernel machine is described in this section. It is a straight forward hardware implementation of the MP-based mathematical  formulation proposed in section \ref{MP_Math}. The salient features of the proposed design are as follows:
\begin{itemize}
  \item The proposed architecture is designed to support 32 input features and a total of 256 kernels. It can easily be modified to support any number of features and kernel sizes by increasing or decreasing the MP processing blocks. 
  \item 
  The design parameters are set to support 32 input features (in differential mode $x_{i}^{+}\in \mathbb{R}^{32}$ and $x_{i}^{-}\in \mathbb{R}^{32}$) and 256 stored vectors (in differential form $x_{si}^{+} \in \mathbb{R}^{32}$ and $x_{si}^{-}\in \mathbb{R}^{32}$). Here, $x_{i}^{+}, x_{i}^{-} \in \x,$ and $x_{si}^{+} , x_{si}^{-} \in \x_s$.
  \item In this design, the width of the data path is set to 12-bits. 
  \item The proposed design includes inference and training and does not consume any complex arithmetic modules like multiplier or divider, only uses adder, subtractor, and shifter modules.
  \item The resource sharing between training and inference modules saves a significant amount of hardware resources. The weight vectors (also in differential form $\w^{+}$ and $\w^{-}$) calculation, as shown in red in Fig.\ref{top_mpsvm}, are the additional blocks required for training.  

\end{itemize}

\begin{figure}[!ht]
\centerline{\includegraphics[page=1,scale=0.7,trim=0 0 0 0,clip]{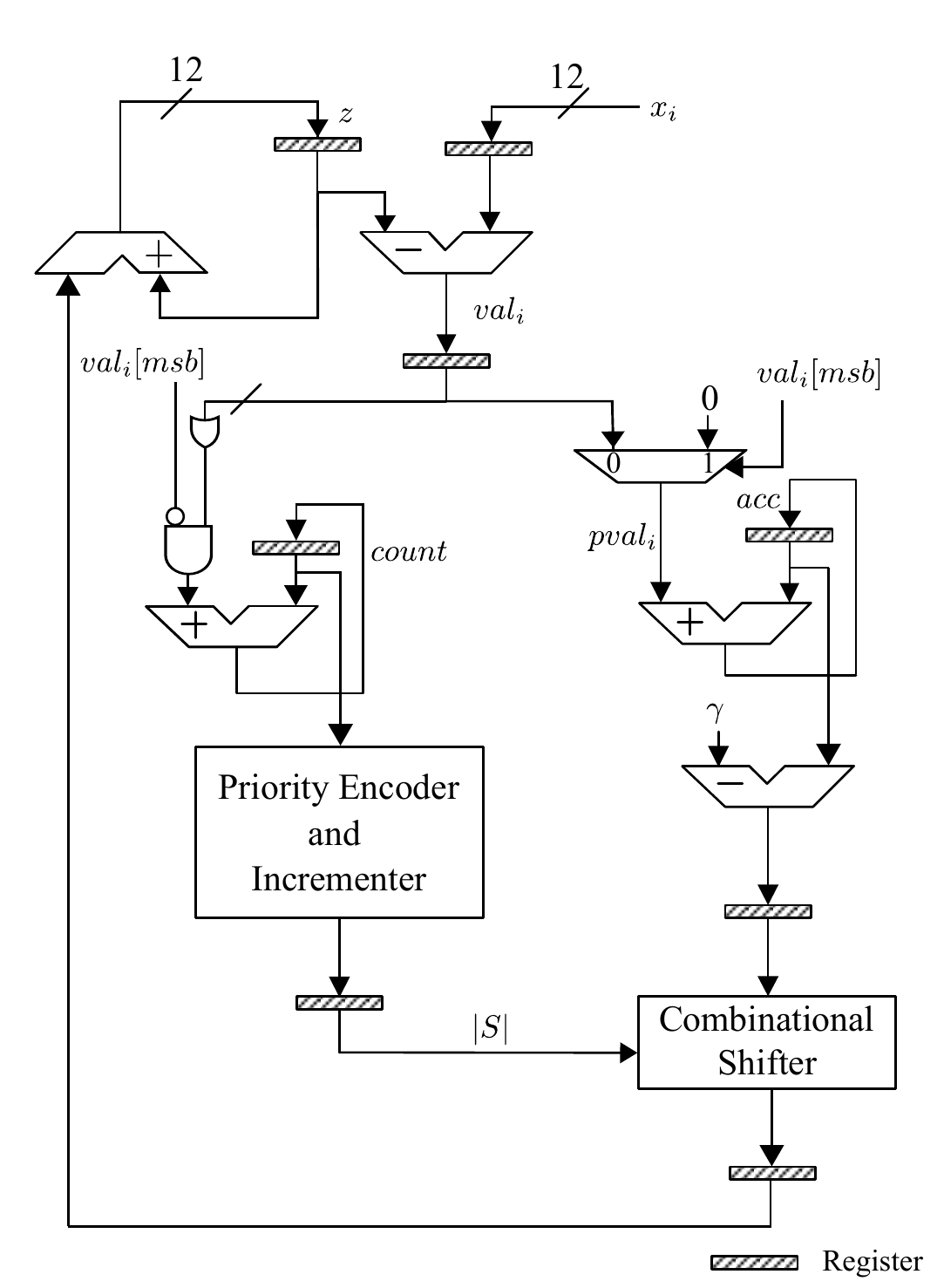}}
\caption{ Architecture of an MP processing block.}
\label{mpfunction}
\end{figure}

\begin{figure}[!ht]
\centerline{\includegraphics[page=1,scale=0.7,trim=0 0 0 0,clip]{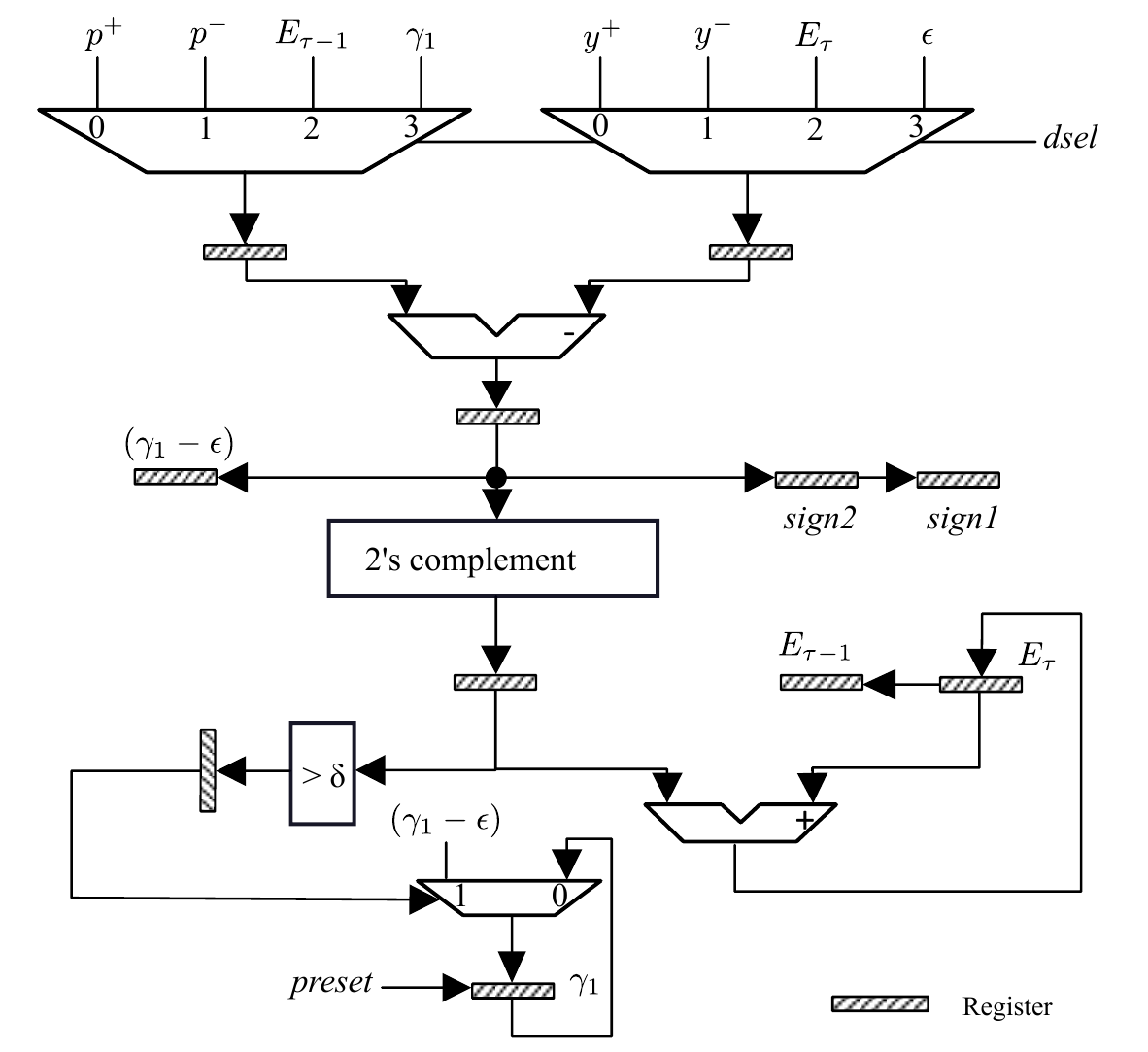}}
\caption{ Architecture for cost function calculation and gamma annealing.}
\label{gammacor}
\end{figure}

\begin{figure*}[ht]
\centerline{\includegraphics[page=1,scale=0.7,trim=0 0 0 0,clip]{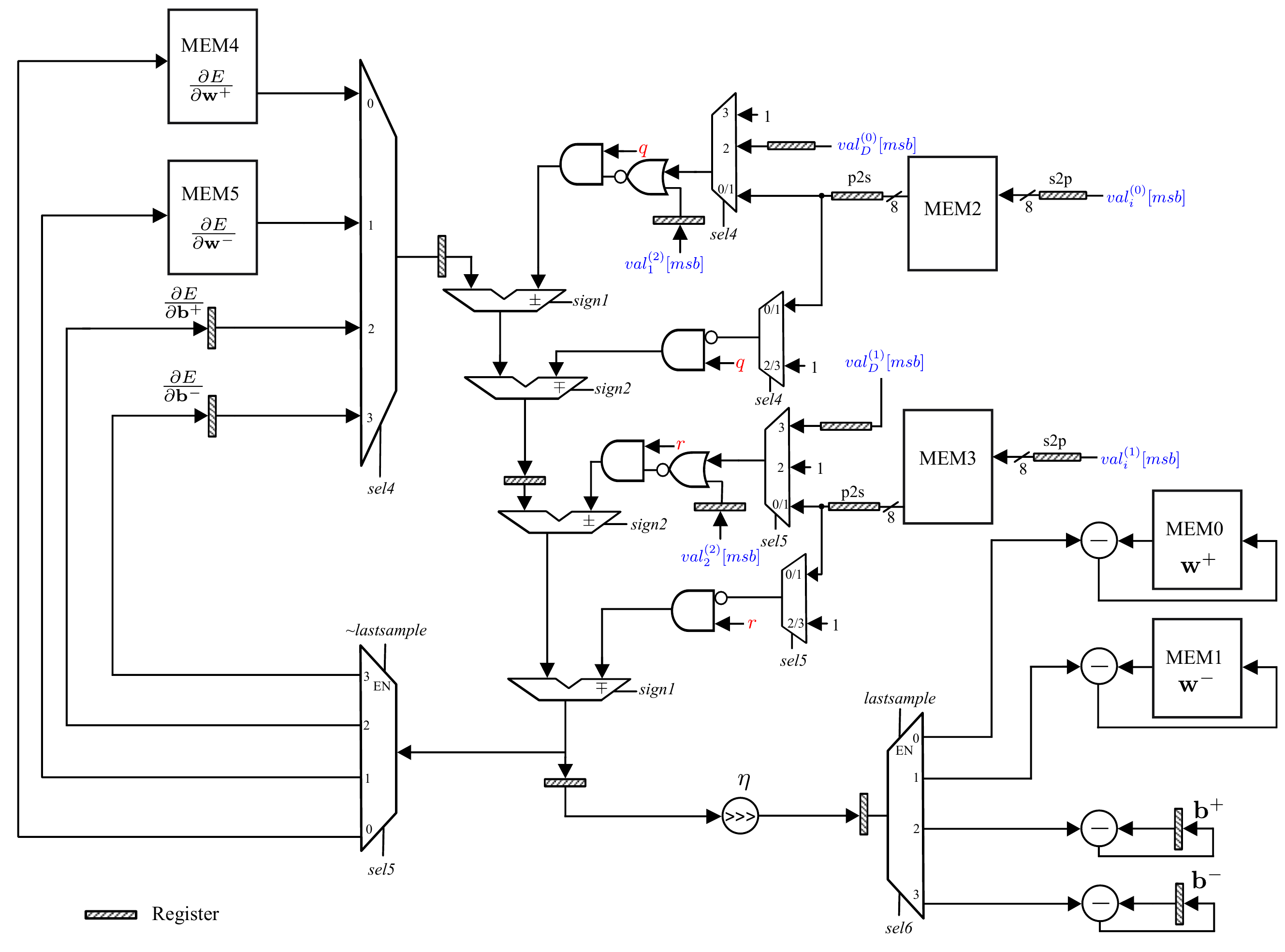}}
\caption{ Architecture for weights and bias update module. The final weights ($\w^{+}$, $\w^{-}$) and bias values ($\mathbf{b^+}$, $\mathbf{b^-}$) are stored in the corresponding memory blocks (MEM0 and MEM1) and registers, respectively. The parameters $q=\left(1 - \frac{1}{|S|}\right)\frac{1}{|S_p|}$ and $r=\left(1 - \frac{1}{|S|}\right)\frac{1}{|S_n|}$ (shown in red) are calculated using the adder and combinational shifter modules. The $val_i$ (shown in blue) values are provided by the MP processing block Fig.\ref{mpfunction}}
\label{wt2}
\end{figure*} 


\subsection{Description of the proposed design:}
The high-level block diagram of the proposed MP-kernel machine, which can support both online training and inference, is shown in Fig.\ref{top_mpsvm}. The architecture has four stages. Stage 0 shows a memory management unit for storing input samples from the external world to input memory blocks. The MP-based kernel function computation is described in Stage 1. Stage 2 provides a forward pass to generate necessary outputs for inference and training based on the kernel function output obtained from Stage 1. Stage 3 is used only for online training and calculates weight vectors ($\w^{+}$,$\w^{-}$) and bias values ($\mathbf{b^+}$, $\mathbf{b^-}$). Here all the stages execute in a parallel manner after receiving the desired data and control signals. 

\textbf{ Stage 0 (Accessing inputs from the external world)}: The input features ($x_{i}^{+}$, $x_{i}^{-} \in \x$ ) of a sample are stored in Block RAM (BRAM), either IPMEM0 or IPMEM1. These BRAMs act as a ping-pong buffer. When an incoming sample is being written into IPMEM0, the kernel computation is carried out on the previously acquired sample in IPMEM1 and vice-versa. The dimension of each input memory block is $64\times9$-bit. 


\textbf{Stage 1 (Kernel function computation):} 
The architectural design of an MP processing block is shown in Fig.\ref{mpfunction}, the straight forward implementation of the algorithm from Fig. \ref{algo:NR}. The inputs $x_{i}$ appear in the MP unit serially i.e. one input in each clock cycle, and calculates $val_i$. The $val_i$ having positive value is getting accumulated in $acc$ register, where as number of positive terms is counted on $count$ register. The $msb$ of $val_i$ is used to detect positive terms. The $val_i$ is passed through the OR gate to perform bit-wise OR and the output is used to discard $val_i$ with value zero. After accessing all the inputs, the $count$ value is approximated to nearest upper bound of power of two, i.e. $|S|\approx 2^{P}$ and $P = floor(log_2(count)) + 1$. This is implemented using a simple priority encoder followed by an incrementer. The priority encoder checks the highest bit location whose bit value is $1$ for the variable $count$.
The combinational shifter right shifts the $(acc-\gamma)$ value by $P$ number of bits. After that, $z$ is updated with the summation of the previous value of $z$ and shifter output. This process iterates 10 times to get the final value of $z$.
The high-level architecture for computing MP kernel function ($\K^{-}$) represented in eq.\eqref{eq:15}, is shown in Fig.\ref{top_mpsvm}. Here, 64 MP processing blocks (KMP0-63) are executed parallely and reuses these blocks 4 times to generate a kernel vector ($\K^{-}$) of dimension $256\times1$. Each MP processing block is associated with a dual-port BRAM named SVMEM for storing 4 support vectors. Here, SVMEM0 for KMP0  stores support vectors with indices 0, 64, 128, and 192. Similarly, SVMEM1 stores the vectors $x_{si}$ with indices 1, 65, 129, and 193. The rest of the support vectors are stored similarly in the respective SVMEM. An MP processing block receives the inputs serially as mentioned in eq.\eqref{eq:15}, i.e., $2x
_{i}^{+}$, $2x_{si}^{-}$, $x_{si}^{-}+x_{i}^{+}+2$, $2x_{i}^{-}$, $2x_{si}^{+}$, $x_{si}^{+}+x_{i}^{-}+2$ and generates output $z$ after 10 iterations. Here, all the 64 MP processing blocks receives inputs and generates outputs simultaneously. The generated kernel vector $\K^-$ is stored in BRAM either KMEM0 or KMEM1 based on the select signal $sel1$ (working as a ping-pong buffer). Here, $\K^{+}$=$-\K^{-}$.

\begin{table*}[ht]
\centering
\caption{Comparison of Architecture and Resource Utilization of Related Work.\label{RLUT}}
 \begin{threeparttable}[t]
\begin{tabular}{|c||c|c|c|c|c|c|c|}
\hline
\textbf{\begin{tabular}[c]{@{}c@{}}\\ Hardware \\ Comparison \end{tabular}} & \textbf{\begin{tabular}[c]{@{}c@{}} Kuan et.al \\ \cite{kuan2011vlsi}\end{tabular}}                                                                                                         & \textbf{\begin{tabular}[c]{@{}c@{}} Peng et.al\\ \cite{peng2013rec}\end{tabular}}                                                                       & \textbf{\begin{tabular}[c]{@{}c@{}}Peng et.al\\ \cite{peng2014trainable}\end{tabular}}                                                                                                   & \textbf{\begin{tabular}[c]{@{}c@{}}Feng et.al\\ \cite{feng2017vlsi}\end{tabular}}                                                                          & 
\textbf{\begin{tabular}[c]{@{}c@{}} Jiang et.al \\ \cite{jiang2017fpga}\end{tabular}} & 
\textbf{\begin{tabular}[c]{@{}c@{}} Mandal et.al \\ \cite{mandal2014implementation}\end{tabular}} & 
\textbf{\begin{tabular}[c]{@{}c@{}}This Work\end{tabular}}  \\ \hline \hline
\textbf{FPGA}                                          &\begin{tabular}[c]{@{}c@{}}Cyclone II\\ DE2-70\end{tabular}                                                                                                                        & \begin{tabular}[c]{@{}c@{}}Spartan-3 \\ c3s4000\end{tabular}                                                                                      & \begin{tabular}[c]{@{}c@{}}Cyclone II \\ 2C70\end{tabular}                                                                                                                  & \begin{tabular}[c]{@{}c@{}}Cyclone II \\ ep2c70f896C6\end{tabular}                       
&\begin{tabular}[c]{@{}c@{}}Zynq \\ 7000\end{tabular}
&\begin{tabular}[c]{@{}c@{}}Virtex \\ 7vx485tffg1157\end{tabular}
&\begin{tabular}[c]{@{}c@{}}Artix-7 \\ xc7a25tcpg238-3\end{tabular}                                                                                                                                   \\ \hline
\textbf{Operating Frequency}                                          & 50 MHz                                                                                                                      & 50 MHz                                                                                     & 50 MHz                                                                                                                  & 50 MHz  & NA\tnote{+} & NA\tnote{+}
& 62 MHz (Max.200 MHz)\tnote{*}                                                                                                                    \\ \hline
\textbf{Multipliers}     & 58                                                                                                                      & NA\tnote{+}                                                                                     & NA\tnote{+}                                                                                                                  & 41                                                                                                                      &152 &515 & 0                                                                                                                    \\ \hline
\textbf{Resources}     & 6395 (LEs)                                                                                                                      & 5612 (LEs)                                                                                     & 5755 (LEs)                                                                                                                  & 6839 (LEs)                                                                                                                     & \begin{tabular}[c]{@{}c@{}}21305 (FFs) \\ 14028\tnote{*} (LUTs)\end{tabular} 
& \begin{tabular}[c]{@{}c@{}}19023 (FFs) \\ 19023 (LUTs)\end{tabular} 
& \begin{tabular}[c]{@{}c@{}}9735 (Max. 9788)\tnote{*} (FFs) \\ 9535 (Max. 9874)\tnote{*} (LUTs)\end{tabular}                                                                                                                    \\ \hline
\textbf{RAM (Kbits)}     & 445                                                                                                                       & 432                                                                                     & 456                                                                                                                  & 304   & NA\tnote{+} & NA\tnote{+}                                                                                                                    & 317                                                                                                                    \\ \hline

\textbf{Input Vector Size}                                                  & 16               
& 16    
& 16 
& 8
& NA\tnote{+}
& 12
& 32                                                                                                                 \\ \hline

\textbf{Vector bit length}                                                  & 24-bit               
& 24-bit    
& 24-bit 
& 16-bit
& 25-bit
& NA\tnote{+}
& 12-bit                                                                                                                 \\ \hline

\textbf{Training Cycles}                                                  & 454M               
& 13M    
& 60M 
& 2964K
& NA\tnote{+}
& NA\tnote{+}
& 2054K                                                                                                                 \\ \hline

\begin{tabular}[c]{@{}c@{}}\textbf{On-Chip Classification} \\ \textbf{Support}\end{tabular}                                                  & No               
& No    
& Yes 
& No
& No
& No
& Yes                                                                                                                 \\ \hline

\textbf{Dynamic Power}                                                  & 216 mW               
& 206 mW    
& 195 mW 
& 45 mW
& NA\tnote{+}
& NA\tnote{+}
& 107 mW                                                                                                                 \\ \hline

\textbf{Energy Consumed\tnote{$\dagger$}}                                                  & 116 pJ               
& NA\tnote{+}    
& NA\tnote{+} 
& 45.1 pJ
& NA\tnote{+}
& NA\tnote{+}
& 13.4 pJ                                                                                                                 \\ \hline

\end{tabular}
\begin{tablenotes}
     \item[+] These works did not report the corresponding values for their designs. \\
     \item[*] These values correspond to design changes for operating at a frequency of 200 MHz.\\
     \item[$\dagger$] Based on analysis in \cite{horowitz20141} for energy consumed for different operations on 45 nm technology node. Multipliers and Adders are used as operations in this case. \\
     \item[] Note that LUTs/FFs from Xilinx and LEs from Intel are not equivalent. 
   \end{tablenotes}
   \end{threeparttable}%
\end{table*}

\textbf{Stage 2 (Merging of Kernel function output):} 
In this stage, three MP processing blocks (MMP0-2) are arranged in a layered manner. In the first layer, two MP processing blocks, i.e., MMP0 and MMP1, execute simultaneously, and in the second layer, MMP2 starts processing after receiving outputs from the first layer. The MMP0 implements the eq.\eqref{eq:38} and generates output $z^{+}$. The inputs arrive at the MP block serially in the following order $(\w^{+}$+ $\K^{+})$, $(\w^{-}$+ $\K^{-})$, $\mathbf{b^+}$. Similarly, MMP1 takes ($\w^{+}$+ $\K^{-}$), ($\w^{-}$+ $\K^{+}$), $\mathbf{b^-}$, and $\gamma_1$  as inputs and produces $z^{-}$ as the output according to eq.\eqref{eq:39}.  
In this stage, $\K^{-}$ is accessed from either KMEM0 or KMEM1 based on $sel2$ signal, $\w^{+}$ and $\w^{-}$ are accessed from MEM0 and MEM1, respectively.


The two outputs ($z^+$ and $z^-$) generated from the first layer are provided as inputs to MMP2 along with $\gamma_1$. This module generates outputs $p^{+}$ and $p^{-}$ according to the eq.\eqref{eq:17}. The final prediction value ${p}$ is computed based on the eq.\eqref{eq:16}. In the Fig.\ref{mpfunction} $pval_1$ = $p^+$ and $pval_2$ = $p^-$ at $10^{th}$ round.  

\textbf{Stage 3 (Weights and bias update module):} 
This stage executes only during the training cycle. The detailed hardware architecture for updating weights and bias is shown in Fig.\ref{wt2}. The proposed design performs error gradients update ($\frac{\partial E}{\partial \w^+}$, $\frac{\partial E}{\partial \w^-}$, $\frac{\partial E}{\partial \mathbf{b^+}}$, $\frac{\partial E}{\partial \mathbf{b^-}}$) followed by weights and bias update at the end of each iteration ($\tau$) according to eq.\eqref{eq:33}-\eqref{eq:36}. 
The hardware resources of stages 0, 1 and 2 are shared by both the training and inference cycles.  This stage updates weight vectors ($\w^+$, $\w^-$) and bias values ($\mathbf{b^+}$, $\mathbf{b^-}$) based on the parameters generated at Stage 2. 
At the training phase, outputs $|S_p|$, $|S_n|$ and $|S|$ are generated from MMP0, MMP1, and MMP2, respectively, and stored in three different registers. After that, two parameters $q=\left(1 - \frac{1}{|S|}\right)\frac{1}{|S_p|}$ and $r=\left(1 - \frac{1}{|S|}\right)\frac{1}{|S_n|}$ are calculated using the adder and combinational shifter modules. The precomputed values $q$ and $r$ are used to update the weights and bias values according to eq.\eqref{eq:33}-\eqref{eq:36}. 
The sign bit, i.e., $val_i^{(0)}[msb]$ and $val_i^{(1)}[msb]$, originated from MMP0 and MMP1 respectively at the $10^{th}$ round are passed through two separate 8-bit serial in parallel out (s2p) registers to generate 8-bit data which is stored in two separate BRAMs (MEM2 and MEM3, respectively). The memory contents are accessed through an 8-bit parallel in serial out  (p2s) register during processing. Similarly, the sign bits $val_1^{(2)}[msb]$ and $val_2^{(2)}[msb]$ from MMP2 are also stored in two separate registers. The sign bits are used to implement indicator function ($\mathbbm{1}$) represented in eq.\eqref{eq:33}-\eqref{eq:36}.  
An architecture that calculates both cost function ($E$) and gamma annealing is shown in Fig. \ref{gammacor}. The architecture works in two phases, in the first phase ($dsel[msb]=0$), the cost function is calculated according to eq.\eqref{eq:20}, and in the second phase ($dsel[msb]=1$), gamma annealing is performed as per the Algorithm \ref{algo:gamma} mentioned in Section \ref{On-chip}. The $preset$ signal initializes $\gamma_1$ register and the register content getting updated in each iteration ($\tau$) according to the Algorithm \ref{algo:gamma}. In the first phase, the module also calculates $\sgn(p^+ - y^+)$, $\sgn(p^- - y^-)$ and stores the outputs in two registers ($sign1$ and $sign2$), respectively, which are used as inputs to the weights and bias update module. 
The datapath for updating error gradients as well as weights and bias updates are shown in Fig. \ref{wt2}. The values of $\frac{\partial E}{\partial \w^+}$ and $\frac{\partial E}{\partial \w^-}$ are stored in MEM4 and MEM5 BRAMs (256$\times$9-bit each), respectively, and the values of $\frac{\partial E}{\partial \mathbf{b^+}}$ and $\frac{\partial E}{\partial \mathbf{b^-}}$ are stored in registers.
At the beginning of each iteration ($\tau$), these are initialized with zero. The error gradients are accumulated for each sample, and the respective storage is updated. The mathematical formulation for the error gradient update is explained in Section B of the supplementary document. 

In Fig. \ref{wt2}, the MUXes are used to select appropriate inputs for the adder-subtractor modules. The signals $sign1$ and $sign2$ are used to select the appropriate operations i.e., either addition or subtraction. The $sel5$ signal is generated by delaying $sel4$ by one clock cycle. When the $msb$ of $sel4$ is 0, it accesses and updates the MEM4 and MEM5 alternatively in each clock cycle. After that, the $msb$ of $sel4$ becomes high and updates the registers. 
While processing the last sample of an iteration ($\tau$), an active high signal $lastsample$ along with $sel6$ starts updating the MEM0 and MEM1 for $\w^+$ and $\w^-$, and the registers for $\mathbf{b^+}$ and $\mathbf{b^-}$. Here, the learning rate ($\eta$) is expressed in powers of 2 and can be implemented using a combinational shifter module. 
\section{Results and Discussion} \label{Results}

The proposed design has been implemented on Artix-7 (xc7a25tcpg238-3), manufactured at 28 nm technology node, a low-powered, highly efficient FPGA. Artix-7 family devices have been extensively used in edge device applications like automotive sensor processing. This makes it ideal for us to showcase our design capability on this device. Our design is capable of running at a maximum frequency of 200 MHz. The design supports both inference and on-chip training. It consumes about 9874 LUTs, 9788 FFs, along with 35 BRAMs (36 Kb). The design does not utilize any complex arithmetic module such as multiplier. For processing a sample, the proposed MP-kernel machine consumes 8024 clock cycles for computing a ($256\times1$) kernel vector $\K^{-}$ (Stage 1). Stage 2 consumes 5256 and 5710 clock cycles during inference and learning, respectively. Stage 3, which is active during learning, consumes 524 clock cycles. In top level, the proposed design has two sections : section 1, executes stage 1 and section 2, executes stage 2 and stage 3. Two sections work in pipeline fashion. In real time applications, when pipe is full, it can generate one output after 8024 clock cycles. 


\begin{table}[ht]
\centering
\caption{Comparison of Traditional SVM Inference and MP Kernel Machine (Training and Inference) Resource Utilization at operating frequency of 62 MHz.\label{RLUTHLS}}
\begin{tabular}{|c||c|c|c|}
\hline
\textbf{\begin{tabular}[c]{@{}c@{}}Hardware\\ Resources\end{tabular}} & \textbf{\begin{tabular}[c]{@{}c@{}}Traditional\\ SVM \\ (Inference)\end{tabular}} & \multicolumn{2}{c|}{\textbf{\begin{tabular}[c]{@{}c@{}}MP\\ Kernel Machine\\ (Training and Inference)\end{tabular}}}     \\ \hline \hline
\textbf{FPGA}                                                         & \begin{tabular}[c]{@{}c@{}}xc7z020-1\\ clg400c\end{tabular}                       & \begin{tabular}[c]{@{}c@{}}xc7z020-1\\ clg400c\end{tabular} & \begin{tabular}[c]{@{}c@{}}xc7a25\\ tcpg238-3\end{tabular} \\ \hline
\textbf{FFs}                                                          & 6148                                                                              & 9734                                                        & 9735   \\ \hline
\textbf{LUTs}                                                         & 18141                                                                             & 9572                                                        & 9535                                                       \\ \hline                                                     
\textbf{BRAMs (36k)}                                                   & 46                                                                                & 35                                                          & 35                                                         \\ \hline
\textbf{DSPs}                                                         & 192                                                                               & 0                                                           & 0                                                          \\ \hline
\textbf{Dynamic Power}                                                & 320 mW                                                                            & 107 mW                                                      & 107 mW                                                     \\ \hline
\end{tabular}
\end{table}
We compare our system with similar SVM systems in the literature. From Table \ref{RLUT}, it is clear that our system consumes the least amount of energy of 13.4 pJ compared to similar SVM systems with online training capability. Based on the type and number of operations used in each design and the energy consumed by each type of operator, using \cite{horowitz20141} as the reference, we arrive at the energy consumption of each system. Our system can process an input vector size of 32, which is higher than other systems, and at the same time consumes lower RAM bits with no DSP usage making it resource efficient. The amount of training cycles required by our system is lowest, i.e, around 2054K, providing lower latency and higher throughput. A traditional SVM inference algorithm is also implemented on the PYNQ board (xc7z020clg400-1), manufactured at 28 nm technology node, to compare resource utilization and power consumption with the proposed MP kernel machine algorithm. We used linear kernel for this implementation as non-linear kernel was complex for implementation. The hardware design for the traditional SVM algorithm consumed a higher number of resources, and due to this, we were unable to fit the design in the same FPGA part used for the kernel machine design. To obtain a fair comparison, we implemented our design on the same PYNQ board. The design parameters for both designs are the same to make a direct comparison and bring out the efficiency of our system. The maximum operating frequency of the traditional SVM design is limited to 62 MHz. The power and efficiency of our system, which includes training and inference, is compared to this traditional SVM implementation having only an inference engine in Table \ref{RLUTHLS}. We see that our MP kernel machine consumes a fraction of the power and about half of the LUTs (including the training engine) in comparison to the traditional SVM. Due to the multiplierless design, we see no DSPs consumed compared to 192 DSPs in traditional SVM design.  
For edge devices, the requirement of a small footprint and low power is fulfilled by our system.

\begin{table}[]
\centering
\caption{Accuracy in \% for UCI Datasets in Percent.\label{accuracy}}
 \begin{threeparttable}[t]
\begin{tabular}{|c||c|c|c|c|c|c|}
\hline
\multirow{3}{*}{  \textbf{Datasets}}    & \multicolumn{4}{c|}{\textbf{Full Precision}}                                                                                                                             & \multicolumn{2}{c|}{\textbf{\begin{tabular}[c]{@{}c@{}}\\Fixed Point \\ (12-bit)\end{tabular}}} \\ \cline{2-7} 
& \multicolumn{2}{c|}{\textbf{\begin{tabular}[c]{@{}c@{}}\\Traditional\\ SVM\tnote{*} \end{tabular}}} & \multicolumn{2}{c|}{\textbf{\begin{tabular}[c]{@{}c@{}}\\MP\\ SVM\end{tabular}}} & \multicolumn{2}{c|}{\textbf{\begin{tabular}[c]{@{}c@{}}\\MP\\ SVM\end{tabular}}}                \\ \cline{2-7}

                                                                       & \textbf{\begin{tabular}[c]{@{}c@{}}\\ Train \end{tabular}}                              & \textbf{\begin{tabular}[c]{@{}c@{}}\\ Test \end{tabular}}                             & \textbf{\begin{tabular}[c]{@{}c@{}}\\ Train \end{tabular}}                         & \textbf{\begin{tabular}[c]{@{}c@{}}\\ Test \end{tabular}}                          & \textbf{\begin{tabular}[c]{@{}c@{}}\\ Train \end{tabular}}                                 & \textbf{\begin{tabular}[c]{@{}c@{}}\\ Test \end{tabular}}                                 \\ \hline \hline  
                                                                           
\textbf{\begin{tabular}[c]{@{}c@{}}Occupancy \\ Detection\end{tabular}}  & 98.6 \cite{liu2017intelligent}                                         & 97.8 \cite{liu2017intelligent}                                       & 98                                     & 94                                    & 97.9                                              & 93.8                                            \\ \hline
\textbf{\begin{tabular}[c]{@{}c@{}}FSDD\\ Jackson\end{tabular}} & 99 \cite{nair2021filter}                                         & 99 \cite{nair2021filter}                                        & 97.5                                     & 96                                    & 97                                              & 96                                            \\ \hline
\textbf{\begin{tabular}[c]{@{}c@{}}AReM\\ Bending\end{tabular}}            & 96 \cite{anguita2013public}                                          & 96 \cite{anguita2013public}                                        & 96                                     & 94.1                                    & 96                                              & 93.2                                            \\ \hline
\textbf{\begin{tabular}[c]{@{}c@{}}AReM\\ Lying\end{tabular}}              & 96 \cite{anguita2013public}                                          & 96 \cite{anguita2013public}                                        & 96                                     & 94.7                                    & 95.9                                              & 94                                            \\ \hline
\end{tabular}
\begin{tablenotes}
     \item[*] These values are reported in different works.
   \end{tablenotes}
    \end{threeparttable}[t]
\end{table}

\begin{figure}[ht]
    \centerline{\includegraphics[page=1,scale=0.42,trim=0 0 0 0,clip]{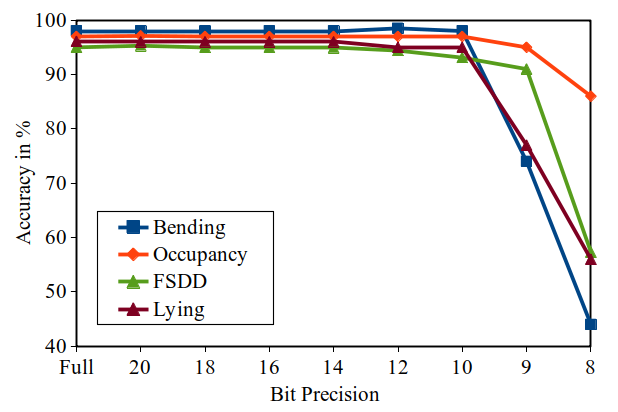}}
        \caption{Bit precision vs. Accuracy for Datasets}
    \label{fig:bitprecisionacc}
\end{figure}

We used datasets from different domains for classification to benchmark our system.  We used the occupancy detection dataset from the UCI ML repository \cite{Dua:2019} \cite{candanedo2016accurate}, which detects whether a particular room is occupied or not based on different sensors like light, temperature, humidity, and $CO_2$.  We also verified our system on the Activity Recognition dataset (AReM) \cite{palumbo2016human}. We used a one versus all approach on the AReM dataset for binary classification using this system. We chose two of the activities as positive cases, i.e., bending and lying activities, to verify the classification the capability of our system. 
Free Spoken Digits Dataset (FSDD) was used to showcase the capability of our system for speech applications \cite{jackson2016free}. Here, we used this dataset for the speaker verification task. We identified a speaker named Jackson among 3 other speakers. Mel-Frequency Cepstral Coefficients (MFCC) is used as a feature extractor for this classification.  


 Since our hardware currently supports 256 input samples, we truncated the datasets to 256 inputs. We performed a k-fold cross-validation to generate 256 separate train and test samples per fold from the original dataset to arrive at the results. This was repeated to cover the entire size of the dataset and get the average accuracy results across separate runs.
A MATLAB model of the proposed architecture is developed to determine the datapath precision. We can see from Fig.\ref{fig:bitprecisionacc}, that the dataset's accuracy remains more or less constant as we reduce the bit width precision up to 12-bit. Below 12-bit, accuracy starts degrading due to quantization error. Hence, we used 12-bit precision for implementing the datapath. The accuracy degradation between full precision MATLAB model and fixed point (12-bit) RTL versions were minimal, as shown in Table \ref{accuracy}. We compared the results of traditional SVM from independent works using the same datasets with our MP-based system. Despite being an approximation, we can get the accuracy results of our system comparable to the traditional SVM systems with minimal degradation in accuracy in some cases. From these results, our system demonstrates its capability in classification tasks, and also, with minor changes to the hardware, it can adapt to any application.


\section{Conclusion} \label{Conclusion}
In this paper, we show a novel algorithm used for classification in edge devices using MP function approximation. This algorithm proves to be hardware-friendly since the training and inference algorithm can be implemented using basic primitives like adders, comparators, counters, and threshold operations. The unique training formulation for kernel machines is lightweight and hence enables online training. The same hardware is used for training and inference, leveraging hardware reuse policy to make it a highly efficient system. Also, the system is highly scalable without requiring significant hardware changes. 
In comparison to traditional SVMs, we were able to achieve low power and computational resource usage, making it ideal for edge device deployment. This algorithm being multiplierless improves the speed of operation when compared to traditional SVMs. Such edge devices can be deployed in remote locations for surveillance and medical applications, where human intervention may be minimal. We can fabricate this system into Application Specific Integrated Chip to make it more power and area efficient. Moreover, this algorithm can be extended to more complex ML systems like deep learning to leverage the low hardware footprint.

\ifCLASSOPTIONcaptionsoff
  \newpage
\fi



%
\medskip

\section*{Acknowledgment}
This research was supported in part by SPARC grant (SPARC/2018-2019/P606/SL) and SRAship grant from CSIR-HRDG, and IMPRINT Grant (IMP/2018/000550) from the Department of Science 
and Technology, India. 
The authors would like to acknowledge the joint Memorandum of Understanding (MoU) between Indian Institute of Science, Bangalore and Washington University in St. Louis for supporting this research activity.

\bibliographystyle{IEEEtran}
\bibliography{references}

\end{document}


%
\title{Multiplierless MP-Kernel Machine For Energy-efficient Edge Devices}
%
%
%

\author{Abhishek~Ramdas~Nair$^*$,
        Pallab~Kumar~Nath$^*$,
        Shantanu~Chakrabartty,
        and~Chetan~Singh~Thakur,
\thanks{Abhishek Ramdas Nair, Pallab Kumar Nath and Chetan Singh Thakur are with the Department
of Electronic Systems Engineering, Indian Institute of Science, Bangalore,
KA, 560012 INDIA e-mail: (abhisheknair@iisc.ac.in, pallabkumar@iisc.ac.in, csthakur@iisc.ac.in).}
\thanks{Shantanu Chakrabartty is with Department of Electrical and Systems Engineering, Washington University in St. Louis,USA, 63130. e-mail:shantanu@wustl.edu}
\thanks{$^*$ Both Abhishek Ramdas Nair and Pallab Kumar Nath have contributed equally to this paper.}%
\thanks}

\maketitle

\section{Supplementary Material}
In this document, we present derivations for (A) the Kernel used for MP Kernel Machine in MP domain and (B) Partial Derivatives for MP based Gradient Descent. 
\subsection{Kernel Derivation for MP Kernel Machine} \label{appendix:kernel}
The similarity measure function used as the kernel can be described as,
\begin{align}
    K(\x_s,\x) = \frac{(\x_s - \x)^2}{2} + 2. \label{eq:82} 
\end{align}
Here, $K(\cdot):\mathbb{R}^{d \times d} \rightarrow \mathbb{R}^d$ and $\x_s$, $\x \in  \mathbb{R}^d$.
We use the difference vectors of $\x_s$ and $\x$ to derive for MP domain, i.e., using $\x_s = \x_s^+ - \x_s^-$ and $\x = \x^+ - \x^-$, we can write,
\begin{align}
    (\x_s - \x)^2 = (\x_s^+ - \x_s^- - \x^+ + \x^-)^2. \: \quad \qquad \quad \nonumber \\
                  = \; \x_s^+\x_s^+ + \x_s^-\x_s^- + \x^+\x^+  + \x^-\x^- \nonumber \\
                    - \; 2\x_s^+\x^+ - 2\x_s^+\x_s^- + 2\x_s^+\x^- \quad \quad \: \:\nonumber \\
                    + \; 2\x_s^-\x^+ - 2\x_s^-\x^- - 2\x^+\x^-. \quad \quad \label{eq:86}
\end{align} 
Ensuring $|{\x}| < 1$ or ${\x} = {\x}^{+} - {\x}^{-}$ such that ${\x}^{+} + {\x}^{-} = 1$ and $\x^+, \x^- \geq 0$. Similarly, we get ${\x_s}^{+} + {\x_s}^{-} = 1$. Using these constraints, we can show,
\begin{align}
    4 = (\x_s^+ + \x_s^-)^2 + (\x^+ + \x^-)^2 + 2(\x_s^+ + \x_s^-)(\x^+ + \x^-). \nonumber \\
      = \; \x_s^+\x_s^+ + \x_s^-\x_s^- + \x^+\x^+  + \x^-\x^- \; \; \; \qquad \qquad \qquad \quad \nonumber \\
                    + \; 2\x_s^+\x^+ + 2\x_s^+\x_s^- + 2\x_s^+\x^- \; \; \; \qquad \qquad \qquad \quad \quad \quad \: \:\nonumber \\
                    + \; 2\x_s^-\x^+ + 2\x_s^-\x^- + 2\x^+\x^-. \; \; \: \: \quad \qquad \qquad \qquad  \quad \quad \label{eq:87}
\end{align} 
Adding eq.\eqref{eq:86} and \eqref{eq:87}, we get,
\begin{align}
    (\x_s - \x)^2 + 4 = 2\x_s^+\x_s^+ + 2\x_s^-\x_s^- + 2\x^+\x^+  + 2\x^-\x^- \nonumber \\
                        + \; 4\x_s^+\x^- +  4\x_s^-\x^+. \: \qquad \qquad \qquad \quad \quad  \nonumber \\
    \frac{(\x_s - \x)^2}{2} + 2 = \x_s^+\x_s^+ + \x_s^-\x_s^- + \x^+\x^+  + \x^-\x^- \; \; \: \quad \nonumber \\
                        + \; 2\x_s^+\x^- +  2\x_s^-\x^+. \label{eq:87a} \qquad \qquad \qquad \qquad                    
\end{align}
Thus, using eq.\eqref{eq:82}, we can write,
\begin{align}
    K(\x_s,\x) = \x_s^+\x_s^+ + \x_s^-\x_s^- + \x^+\x^+  + \x^-\x^- \nonumber \\
                        + \; 2\x_s^+\x^- +  2\x_s^-\x^+.  \qquad \qquad \quad \quad\label{eq:88}
\end{align}
Using eq.\eqref{eq:45}, and applying MP approximation based on eq.\eqref{eq:68}, we can express eq.\eqref{eq:88} as,
\begin{align}
    \K^- = MP \: ( [2\x_{s}^{+} , 2\x_{s}^{-}, 2\x^{+}, 2\x^-, \quad \quad \nonumber \\
                \x_{s}^+ + \x^- + 2, \; \; \qquad \qquad \nonumber \\
                \x_{s}^- + \x^+ + 2 ], \gamma_2 ). \; \; \quad \quad 
\end{align}
$\gamma_2$ is the MP hyper-parameter for kernel.

\subsection{Partial Derivatives for MP based Gradient Descent} \label{appendix:partial_derivatives}

Taking partial derivative of eq.\eqref{eq:20} with respect to weights $\w^+$ and $\w^-$ and biases $\mathbf{b^+}$ and $\mathbf{b^-}$, we get,
\begin{align}
    \frac{\partial E}{\partial \w^{+}} = \sum\limits_{n=1}^M \left(\sgn{(p_n^+ - y_n^+)}\frac{\partial p_n^+}{\partial \w^+} + \sgn{(p_n^- - y_n^-)}\frac{\partial p_n^-}{\partial \w^+}\right). \label{eq:21}
\end{align}
\begin{align}
    \frac{\partial E}{\partial \w^{-}} = \sum\limits_{n=1}^M  \left(\sgn{(p_n^+ - y_n^+)}\frac{\partial p_n^+}{\partial \w^-} + \sgn{(p_n^- - y_n^-)}\frac{\partial p_n^-}{\partial \w^-}\right). \label{eq:22}
\end{align}
\begin{align}
    \frac{\partial E}{\partial \mathbf{b^+}} = \sum\limits_{n=1}^M \sgn{(p_n^+ - y_n^+)}\frac{\partial p^+}{\partial \mathbf{b^+}}. \label{eq:29}    
\end{align}
\begin{align}
    \frac{\partial E}{\partial \mathbf{b^-}} = \sum\limits_{n=1}^M \sgn{(p_n^- - y_n^-)}\frac{\partial p_n^-}{\partial \mathbf{b^-}}. \label{eq:30}    
\end{align}
Taking partial derivative of eq.\eqref{eq:55} with respect to $x_i$,
\begin{align}
    \frac{\partial z}{\partial x_i} = \frac{1}{|S|}\mathbbm{1}{(x_i > z)}. \label{eq:23}
\end{align}
where $|S|$ indicates the number of $x_i$ such that $x_i > z$ and $\mathbbm{1}$ is the indicator function. 

The cost function derivatives are obtained by estimating the error in output and taking derivative with respect to the parameters like weights $\w^+, \w^-$ and biases $\mathbf{b^+}, \mathbf{b^-}$. 
Taking partial derivative of $p_n^{+}$ with respect to $\w^+$ and using chain rule with results from eq. \eqref{eq:16} and \eqref{eq:17}, we get, 
\begin{align}
    \frac{\partial p_n^+}{\partial \w^+} = \frac{\partial p_n^+}{\partial z^+} 
                                        \frac{\partial z^+}{\partial(\w^{+} + \K^{+})}
                                        \frac{\partial (\w^{+} + \K^{+})}{\partial \w^+} \nonumber \\ 
                                        + \frac{\partial p_n^+}{\partial z} 
                                        \frac{\partial z}{\partial(\w^{+} + \K^{+})}
                                        \frac{\partial (\w^{+} + \K^{+})}{\partial \w^+}.\label{eq:57}   
\end{align}
Here, $p_n^{+} = f(z^+,z)$.
Since, $z = MP(z^+,z^-)$ and   $\frac{\partial (\w^{+} + \K^{+})}{\partial \w^+} = 1 $,
\begin{align}
    \frac{\partial p_n^+}{\partial \w^+} = \frac{\partial p_n^+}{\partial z^+} 
                                        \frac{\partial z^+}{\partial(\w^{+} + \K^{+})} \nonumber \\ 
                                        + \frac{\partial p_n^+}{\partial z} 
                                        \frac{\partial z}{\partial z^+}
                                        \frac{\partial z^+}{\partial(\w^{+} + \K^{+})}  \nonumber \\ 
                                        + \frac{\partial p_n^+}{\partial z} 
                                        \frac{\partial z}{\partial z^-}
                                        \frac{\partial z^-}{\partial(\w^{+} + \K^{-})}. \label{eq:83}
\end{align}
From eq.\eqref{eq:17} and \eqref{eq:23}, we get,
\begin{align}
    \frac{\partial p_n^+}{\partial z^+} = \mathbbm{1}{(z^+ > z)}. \; \; \: \nonumber \\
    \frac{\partial p_n^+}{\partial z} = - \mathbbm{1}{(z^+ > z)}. \label{eq:58}
\end{align}
Similarly,
\begin{align}
    \frac{\partial z}{\partial z^+} = \frac{1}{|S|}\mathbbm{1}{(z^+ > z)}. \nonumber \\
    \frac{\partial z}{\partial z^-} = \frac{1}{|S|}\mathbbm{1}{(z^- > z)}. \label{eq:84}
\end{align}
\begin{align}
    \frac{\partial z^+}{\partial (\w^{+} + \K^{+})} 
    = \frac{1}{|S_p|}\mathbbm{1}{(\w^{+} + \K^{+} > z^+)}.  \nonumber \\ 
    \frac{\partial z^-}{\partial (\w^{+} + \K^{-})} 
    = \frac{1}{|S_n|}\mathbbm{1}{(\w^{+} + \K^{-} > z^-)}. \label{eq:59} 
\end{align}
Here, $|S_p|$ is the size of the set $S_p = \{ \w^{+} + \K^{+} ; (\w^{+} + \K^{+}) > z^+ , \w^{-} + \K^{-} ; (\w^{-} + \K^{-}) > z^+ \}$ and 
$|S_n|$ is the size of the set $S_n = \{ \w^{+} + \K^{-} ; (\w^{+} + \K^{-}) > z^- , \w^{-} + \K^{+} ; (\w^{-} + \K^{+}) > z^- \}$. 

Substituting eq.\eqref{eq:58}, \eqref{eq:84} and \eqref{eq:59} in eq.\eqref{eq:83}, 
\begin{align}
    \frac{\partial p_n^+}{\partial \w^+} = \qquad \qquad \qquad \qquad \qquad \nonumber \\
     \mathbbm{1}{(z^+ > z)}\frac{1}{|S_p|}\mathbbm{1}{(\w^{+} + \K^{+} > z^+)} \; \qquad \qquad \qquad \nonumber \\
    - \mathbbm{1}{(z^+ > z)}\frac{1}{|S|}\mathbbm{1}{(z^+ > z)}\frac{1}{|S_p|}\mathbbm{1}{(\w^{+} + \K^{+} > z^+)} \; \nonumber \\
    - \mathbbm{1}{(z^+ > z)}\frac{1}{|S|}\mathbbm{1}{(z^- > z)}\frac{1}{|S_n|}\mathbbm{1}{(\w^{+} + \K^{-} > z^-)}. \label{eq:85a} 
\end{align}

After factorizing, we get,
\begin{align}
    \frac{\partial p_n^+}{\partial \w^+} =  \qquad \qquad \qquad \qquad \qquad \nonumber \\
    \frac{1}{|S_p|}\mathbbm{1}{(\w^{+} + \K^{+} > z^+)}  
    \biggl[\mathbbm{1}{(z^+ > z)} - \frac{1}{|S|}\mathbbm{1}{(z^+ > z)}\biggr]  \nonumber \\
    - \; \frac{1}{|S|}\mathbbm{1}{(z^+ > z)}\mathbbm{1}{(z^- > z)} 
    \frac{1}{|S_n|}\mathbbm{1}{(\w^{+} + \K^{-} > z^-)}. \quad \label{eq:85} 
\end{align}
In this case, we see that $1 \le |S| \le 2$, i.e., $\mathbbm{1}{(z^+ > z)}$ and $\mathbbm{1}{(z^- > z)} \ge 0$. Hence, if either  $\mathbbm{1}{(z^+ > z)}$ or $\mathbbm{1}{(z^- > z)}$ is equal to $0$, then $|S| = 1$, which results in $\frac{\partial p_n^+}{\partial \w^+} = 0$. Thus, we can state that $|S| > 1$.
Using this, we can state $\mathbbm{1}{(z^+ > z)}\mathbbm{1}{(z^- > z)} = |S| - 1$ and rearranging the terms we get,
\begin{align}
   \frac{\partial p_n^+}{\partial \w^+}=   \left(1 - \frac{1}{|S|}\right)\biggl[\mathbbm{1}{(z^+ > z)}\frac{1}{|S_p|}\mathbbm{1}{(\w^{+} + \K^{+} > z^+)} \nonumber \\ 
   - \frac{1}{|S_n|}\mathbbm{1}{(\w^{+} + \K^{-} > z^-)}\biggr].  \quad \qquad \label{eq:25}
\end{align}
Let, $I_{(z^+)}$ = $\mathbbm{1}{(z^+ > z)}$, $I_{(w^+K^+)}$ = $\mathbbm{1}{(\w^{+} + \K^{+} > z^+)}$, $I_{(w^+K^-)}$ = $\mathbbm{1}{(\w^{+} + \K^{-} > z^-)}$,  $I_{\mathbf{(b^+)}}$ = $\mathbbm{1}{(\mathbf{b^+} > z^+)}$ and $I_{\mathbf{(b^-)}}$ = $\mathbbm{1}{(\mathbf{b^-} > z^-)}$,
\begin{align}
   \frac{\partial p_n^+}{\partial \w^+}=   \left(1 - \frac{1}{|S|}\right)\biggl[I_{(z^+)}\frac{1}{|S_p|}I_{(w^+K^+)} - \frac{1}{|S_n|}I_{(w^+K^-)}\biggr].\label{eq:72}  
\end{align}
Similarly, we get values for 
$\frac{\partial p_n^-}{\partial \w^+}, 
\frac{\partial p_n^+}{\partial \w^-},
\frac{\partial p_n^-}{\partial \w^-}, 
\frac{\partial p_n^-}{\partial \mathbf{b^+}}, 
\frac{\partial p_n^+}{\partial \mathbf{b^-}},
\frac{\partial p_n^+}{\partial \mathbf{b^+}}$ and
$\frac{\partial p_n^-}{\partial \mathbf{b^-}}$.
\begin{align}
   \frac{\partial p_n^-}{\partial \w^+}=   \left(1 - \frac{1}{|S|}\right)\biggl[I_{(z^-)}\frac{1}{|S_n|}I_{(w^+K^-)} - \frac{1}{|S_p|}I_{(w^+K^+)}\biggr]. \label{eq:73}  
\end{align}
\begin{align}
   \frac{\partial p_n^+}{\partial \w^-}=   \left(1 - \frac{1}{|S|}\right)\biggl[I_{(z^+)}\frac{1}{|S_p|}I_{(w^-K^-)} - 
   \frac{1}{|S_n|}I_{(w^-K^+)}\biggr]. \label{eq:75}  
\end{align}
\begin{align}
   \frac{\partial p_n^-}{\partial \w^-}=   \left(1 - \frac{1}{|S|}\right)\biggl[I_{(z^-)}\frac{1}{|S_n|}I_{(w^-K^+)} - 
   \frac{1}{|S_p|}I_{(w^-K^-)}\biggr]. \label{eq:76}  
\end{align}
\begin{align}
   \frac{\partial p_n^+}{\partial \mathbf{b^+}}=   \left(1 - \frac{1}{|S|}\right)I_{(z^+)}\frac{1}{|S_p|}I_{\mathbf{(b^+)}}. \label{eq:78}  
\end{align}
\begin{align}
   \frac{\partial p_n^-}{\partial \mathbf{b^-}}=   \left(1 - \frac{1}{|S|}\right)I_{(z^-)}\frac{1}{|S_n|}I_{\mathbf{(b^-)}}. \label{eq:79}  
\end{align}
Substituting the values into the required partial derivative equations,

\begin{align}
    \frac{\partial E}{\partial \w^{+}} = \sum\limits_{n=1}^M \biggl(\sgn{(p_n^+ - y_n^+)}\left(1 - \frac{1}{|S|}\right)\biggl[I_{(z^+)}\frac{1}{|S_p|}I_{(w^+K^+)}  \nonumber \\ - \frac{1}{|S_n|}I_{(w^+K^-)}\biggr] \nonumber \\
    \qquad + \quad \sgn{(p_n^- - y_n^-)}  \left(1 - \frac{1}{|S|}\right)\biggl[I_{(z^-)}\frac{1}{|S_n|}I_{(w^+K^-)} \nonumber \\ - \frac{1}{|S_p|}I_{(w^+K^+)}\biggr] \biggr). \label{eq:74}
\end{align}
\begin{align}
    \frac{\partial E}{\partial \w^{-}} = \sum\limits_{n=1}^M  \biggl(\sgn{(p_n^+ - y_n^+)}\left(1 - \frac{1}{|S|}\right)\biggl[I_{(z^+)}\frac{1}{|S_p|}I_{(w^-K^+)} \nonumber \\  - 
   \frac{1}{|S_n|}I_{(w^-K^-)}\biggr] \nonumber \\ 
    + \quad  \sgn{(p_n^- - y_n^-)}\left(1 - \frac{1}{|S|}\right)\biggl[I_{(z^-)}\frac{1}{|S_n|}I_{(w^-K^-)} \nonumber \\ - 
   \frac{1}{|S_p|}I_{(w^-K^+)}\biggr] \biggr). \label{eq:77}
\end{align}
\begin{align}
    \frac{\partial E}{\partial \mathbf{b^+}} = \sum\limits_{n=1}^M \sgn{(p_n^+ - y_n^+)}\left(1 - \frac{1}{|S|}\right)I_{(z^+)}\frac{1}{|S_p|}I_{\mathbf{(b^+)}}. \label{eq:80}    
\end{align}
\begin{align}
    \frac{\partial E}{\partial \mathbf{b^-}} = \sum\limits_{n=1}^M \sgn{(p_n^- - y_n^-)}\left(1 - \frac{1}{|S|}\right)I_{(z^-)}\frac{1}{|S_n|}I_{\mathbf{(b^-)}}. \label{eq:81}    
\end{align}
